\documentclass{article}
\usepackage{amssymb}
\usepackage{times}
\usepackage{graphicx}
\usepackage{subfigure}
\usepackage{multirow}
\usepackage{multicol}
\usepackage{amsmath}
\usepackage{algorithm}
\usepackage{algorithmic}
\usepackage{threeparttable}
\usepackage{dcolumn}
\usepackage{multirow}
\usepackage{multicol}
\usepackage{booktabs}
\usepackage{subfigure}
\usepackage{hyperref}  
\usepackage{amsthm}
\usepackage{latexsym}
\usepackage{rotating}
\usepackage{textcomp}

\begin{document}
\title{An Adaptive Framework to Tune the Coordinate Systems in Evolutionary Algorithms}
\author{Zhi-Zhong Liu,
        Yong Wang,
        Shengxiang Yang, and Ke Tang
\thanks{Zhi-Zhong Liu is with the School of Information Science and Engineering, Central South University, Changsha 410083, China (Email: zhizhongliu@csu.edu.cn)}
\thanks{Y. Wang is with the School of Information Science and Engineering, Central South University, Changsha 410083, China, and also with the Centre for Computational Intelligence (CCI), School of Computer Science and Informatics, De Montfort University, Leicester LE1 9BH, UK. (Email: ywang@csu.edu.cn)}
\thanks{S. Yang is with the Centre for Computational Intelligence (CCI), School of Computer Science and Informatics, De Montfort University, Leicester LE1 9BH, UK. (Email: syang@dmu.ac.uk)}
\thanks{K. Tang is with Department of Computer Science and Engineering Southern University of Science and Technology, Shenzhen 518055, China. (Email: ketang@ustc.edu.cn)}
}
\maketitle

\begin{abstract}
In the evolutionary computation research community, the performance of most evolutionary algorithms (EAs) depends strongly on their implemented coordinate system. However, the commonly used coordinate system is fixed and not well suited for different function landscapes, EAs thus might not search efficiently. To overcome this shortcoming, in this paper we propose a framework, named ACoS, to $a$daptively tune the ${co}$ordinate $s$ystems in EAs. In ACoS, an Eigen coordinate system is established by making use of the cumulative population distribution information, which can be obtained based on a covariance matrix adaptation strategy and an additional archiving mechanism. Since the population distribution information can reflect the features of the function landscape to some extent, EAs in the Eigen coordinate system have the capability to identify the modality of the function landscape. In addition, the Eigen coordinate system is coupled with the original coordinate system, and they are selected according to a probability vector. The probability vector aims to determine the selection ratio of each coordinate system for each individual, and is adaptively updated based on the collected information from the offspring. ACoS has been applied to two of the most popular EA paradigms, i.e., particle swarm optimization (PSO) and differential evolution (DE), for solving 30 test functions with 30 and 50 dimensions at the 2014 IEEE Congress on Evolutionary Computation. The experimental studies demonstrate its effectiveness.
\end{abstract}


\section{Introduction}\label{sec:Intro}
Evolutionary algorithms (EAs) are a class of population-based meta-heuristic algorithms inspired by biological evolution. EAs exploit bio-inspired mechanisms such as reproduction, mutation, crossover, and selection to evolve a population of candidate solutions toward the optimal solution. Up to now, numerous EA paradigms, such as evolutionary programming (EP) ~\cite{fogel1966artificial}, evolution strategy (ES) ~\cite{rechenberg1978evolutionsstrategien}, genetic algorithm (GA) ~\cite{holland1992adaptation}, genetic programming (GP) ~\cite{Koza1992Genetic}, differential evolution (DE) ~\cite{storn1997differential}, and particle swarm optimization (PSO) ~\cite{eberhart1995new}, have been proposed. Compared with  other types of optimization methods, EAs have some advantages such as ease of use, simple structure, efficiency, and robustness. Therefore, EAs have been broadly applied to diverse fields such as art ~\cite{romero2007art}, economics ~\cite{liao2011novel}, route planning ~\cite{rahat2015hybrid}, robotics ~\cite{silva2015odneat}, and graphic processing ~\cite{Nguyen2016Understanding}.

For most EAs, their performance relies crucially on their implemented coordinate system. However, the original coordinate system, which is the most frequently used coordinate system in current EAs, is fixed throughout the evolutionary process. Under this condition, EAs may fail to produce promising solutions matching the requirements of different function landscapes or even one function landscape at different evolutionary stages. As a result, it is difficult for EAs to search efficiently in the original coordinate system.

To remedy this issue, in some variants of ES and DE, the Eigen coordinate system is established by making use of the population
distribution information. Since the population distribution information can reflect the features of the function landscape to a certain degree, EAs implemented in the Eigen coordinate system thus possess the capability to identify the modality of the function landscape and search efficiently. In 2001, a famous ES called CMA-ES was proposed by Hansen and Ostermeier ~\cite{hansen2001completely}. In CMA-ES, an Eigen coordinate system is established by utilizing the cumulative population distribution information (i.e., the current and historical population distribution information). Afterward, the offspring population is sampled from this Eigen coordinate system. Overall, CMA-ES shows very fast convergence speed and is significantly superior to the ordinary ES. Recently, three attempts (i.e., DE/eig ~\cite{guo2015enhancing} , CoBiDE ~\cite{wang2014differential}, and CPI-DE ~\cite{wang2016utilizing}) have been made to enhance DE's performance by implementing the crossover operator in both the Eigen coordinate system and the original coordinate system. In DE/eig and CoBiDE, only the current population distribution information is extracted to establish the Eigen coordinate system, while like CMA-ES, in CPI-DE the cumulative population distribution information is employed to construct the Eigen coordinate system.
It is interesting to note that all these three attempts in DE draw the similar conclusions: 1) each coordinate system has its own advantages and is suitable for certain kinds of optimization problems, and 2) combining these two coordinate systems can obtain better performance than just using one of them during the evolution. The above conclusions motivate us to design an adaptive scheme to make full use of these two coordinate systems.

This paper presents an adaptive framework, called ACoS, to tune the coordinate systems in EAs. ACoS takes advantage of a covariance matrix adaptation strategy and an additional archiving mechanism to extract cumulative population distribution information, with the aim of establishing the Eigen coordinate system. Moreover, this Eigen coordinate system is synthesized with the original coordinate system, and they are selected based on a probability vector. This probability vector determines the selection ratio of each coordinate system for each individual and is adaptively updated according to the collected information from the offspring. ACoS has been applied to two of the most popular EA paradigms: PSO and DE. The effectiveness of ACoS has been validated by comprehensive experimental studies on $30$ test functions with $30$ and $50$ dimensions at the $2014$ IEEE Congress on Evolutionary Computation (IEEE CEC2014) ~\cite{liang2013problem}.

The main contributions of this paper are summarized as follows:
\begin{itemize}
\item This paper provides a new point of view toward how to describe an evolutionary operator in the original coordinate system, and also offers a convenient transformation from an evolutionary operator in the original coordinate system to the corresponding evolutionary operator in the Eigen coordinate system.
\item A simple yet effective approach is proposed to establish the Eigen coordinate system, which consists of two main elements, i.e., a covariance matrix adaptation strategy and an additional archiving mechanism. In comparison with the previous methods, the cumulative population distribution information derived from our approach is more sufficient.
\item By using a probability vector, this paper presents an adaptive scheme to select an appropriate coordinate system from the original coordinate system and the Eigen coordinate system for each individual during the evolution.
\item The proposed framework (i.e., ACoS) can be readily applied to various EAs. In this paper, we have verified that ACoS is able to improve the performance of two of the most popular EA paradigms: PSO and DE. To the best of our knowledge, it is the first attempt to improve PSO's performance by adjusting the coordinate systems in an adaptive fashion.
\end{itemize}

The rest of this paper is organized as follows. Section~\ref{sec:PSO-DE} briefly introduces PSO and DE. Section~\ref{sec:CoSRW} presents the coordinate systems and their related work. The proposed ACoS and its implementation details are presented in Section~\ref{sec:PA}. The experimental results and the performance comparisons are given in Section~\ref{sec:ES}. Finally, Section~\ref{sec:Conclusion} concludes this paper.

\section{Particle Swarm Optimization (PSO) and Differential Evolution (DE)}\label{sec:PSO-DE}
PSO and DE have become two of the most popular EA paradigms. In this section, we will briefly introduce them.

\subsection{Particle Swarm Optimization (PSO)}\label{sec:PSO}
PSO ~\cite{eberhart1995new} is a population-based stochastic search technique inspired by swarm behavior. It searches with a population (called swarm) of candidate solutions (called particles or individuals). Each particle moves around the search space to seek the global optimum, and its movement is guided by its own personal historical best experience as well as the entire swarm's best experience. Due to ease of use and efficiency, PSO has been successful applied to a variety of real-world optimization problems.

PSO contains two core equations: the velocity updating equation and the position updating equation. At generation $g$, PSO updates the $d$th dimension of the $i$th particle's velocity  $\vec{v}_{i}^{g}= [ v_{i,1}^{g},v_{i,2}^{g},...,v_{i,D}^{g}]^{T} $ and position  $\vec{x}_{i}^{g}= [ x_{i,1}^{g},x_{i,2}^{g},...,x_{i,D}^{g}]^{T} $ as follows:
\begin{equation}\label{eqn:psov}
v_{i,d}^{g+1}=v_{i,d}^{g}+c_{1} r_{1,d}(P_{\text{best}\emph{i},d}^{g}-x_{i,d}^{g})+c_{2} r_{2,d}(G_{\text{best},d}^{g}-x_{i,d}^{g})
\end{equation}
\begin{equation}\label{eqn:psop}
x_{i,d}^{g+1}=v_{i,d}^{g+1}+x_{i,d}^{g}
\end{equation}
where $i\in \{ 1, \ldots ,NP\}$, $d\in \{ 1, \ldots ,D\}$, $NP$ is the population size, $D$ is the dimension of the search space, $\vec{P}_{\text{best}i}^{g}= [P_{\text{best}\emph{i},1}^{g},P_{\text{best}i,2}^{g},...,P_{\text{best}i,D}^{g}]^{T}$ denotes the $i$th particle's historical best position, $\vec{G}_{\text{best}}^{g}= [G_{\text{best},1}^{g},G_{\text{best},2}^{g},...,G_{\text{best},D}^{g}]^{T}$ means the entire swarm's best position, $c_{1}$ and $c_{2}$ are the acceleration parameters, and $r_{1,d}$ and $r_{2,d}$ refer to two uniformly distributed random numbers between $0$ and $1$.

From Eq.\eqref{eqn:psov} and Eq.\eqref{eqn:psop}, it is apparent that PSO works dimension by dimension. Based on the updating of each dimension, the whole velocity  and position of a particle are updated as follows:
\begin{equation}\label{eqn:psovw}
\vec{v}_{i}^{g+1}=\vec{v}_{i}^{g}+c_{1}\textbf{R}_{1}(\vec{P}_{\text{best}\emph{i}}^{g}-\vec{x}_{i}^{g})+c_{2}\textbf{R}_{2} (\vec{G}_{\text{best}}^{g}-\vec{x}_{i}^{g})
\end{equation}
\begin{equation}\label{eqn:psopw}
\vec{x}_{i}^{g+1}=\vec{v}_{i}^{g+1}+\vec{x}_{i}^{g}
\end{equation}
where $\textbf{R}_{1}=diag(r_{1,1},r_{1,2},\ldots, r_{1,D})$ and $\textbf{R}_{2}=diag(r_{2,1},r_{2,2},\ldots, r_{2,D})$.

Since PSO's inception, many researchers have improved its performance in different ways. One way is to control or adjust the particle's velocity. Shi and Eberhart ~\cite{shi1998modified} incorporated an inertial weight $w$ into the original PSO's velocity updating, and Eq.\eqref{eqn:psovw} is thus modified into Eq.\eqref{eqn:psovww}
\begin{equation}\label{eqn:psovww}
\vec{v}_{i}^{g+1}=w\vec{v}_{i}^{g}+c_{1}\textbf{R}_{1}(\vec{P}_{\text{best}i}^{g}-\vec{x}_{i}^{g})+c_{2}\textbf{R}_{2} (\vec{G}_{\text{best}}^{g}-\vec{x}_{i}^{g})
\end{equation}
The only difference between Eq.\eqref{eqn:psovw} and Eq.\eqref{eqn:psovww} is that in Eq.\eqref{eqn:psovww} $w$ is attached to $\vec{v}_{i}^{g}$. In Eq.\eqref{eqn:psovww}, the value of $w$ decreases linearly from 0.9 to 0.4 over the course of search. It is because a larger $w$ in the early stage of evolution is beneficial to exploration, and a smaller $w$ in the later stage of evolution can facilitate the exploitation. In addition, Clerc and Kennedy ~\cite{clerc2002particle} introduced a constriction factor $\chi$ to reformulate the particle's velocity updating:
\begin{equation}\label{eqn:psovwcf}
\vec{v}_{i}^{g+1}=\chi[\vec{v}_{i}^{g}+c_{1}\textbf{R}_{1}(\vec{P}_{\text{best}i}^{g}-\vec{x}_{i}^{g})+c_{2}\textbf{R}_{2} (\vec{G}_{\text{best}}^{g}-\vec{x}_{i}^{g})]
\end{equation}
where $\chi=2/|2-\varphi-\sqrt{\varphi^{2}-4\varphi}|$ and $\varphi=c_{1}+c_{2}$. $\chi$ is preferably set to 0.729 together with $c_{1}=c_{2}=2.05$. For the sake of convenience, the PSO variants in ~\cite{shi1998modified} and ~\cite{clerc2002particle} are called PSO-w and PSO-cf in this paper, respectively, which are two of the most popular PSO variants.

\subsection{Differential Evolution (DE)}\label{sec:DE}
DE ~\cite{storn1997differential} is another simple yet efficient EA paradigm which has been successfully used to deal with a wide spectrum of optimization problems ~\cite{das2011differential}. Similar to other EAs, DE searches with a population of $NP$ individuals: $\textbf{P}^{g}=\{\vec{x}_{i}^{g}=[x_{i,1}^{g},x_{i,2}^{g},...x_{i,D}^{g}]^{T},i=1,2,...,NP\}$, where $g$ denotes the generation number, $NP$ means the population size, and $D$ refers to the dimension of the search space. In DE, at generation $g=0$, the initial population $\textbf{P}^{0}$ is randomly sampled from the search space. After initialization, DE employs mutation, crossover, and selection operators to steer the population toward the global optimum.

\emph{Mutation:} The aim of the mutation operator is to generate a mutant vector  $\vec{v}_{i}^{g}$ for each individual $\vec{x}_{i}^{g}$ (also called a target vector). The following are four commonly used mutation operators in the literature:
\begin{itemize}
\item DE/rand/1
\begin{equation}\label{eqn:derand1}
\vec{v}_{i}^{g}=\vec{x}_{r_{1}}^{g}+F \times (\vec{x}_{r_{2}}^{g}-\vec{x}_{r_{3}}^{g})
\end{equation}
\item DE/rand/2
\begin{equation}\label{eqn:derand2}
\vec{v}_{i}^{g}=\vec{x}_{r_{1}}^{g}+F \times (\vec{x}_{r_{2}}^{g}-\vec{x}_{r_{3}}^{g})+
F \times (\vec{x}_{r_{4}}^{g}-\vec{x}_{r_{5}}^{g})
\end{equation}
\item DE/current-to-best/1
\begin{equation}\label{eqn:decurrenttobest1}
\vec{v}_{i}^{g+}=\vec{x}_{i}^{g}+F \times (\vec{x}_{best}^{g}-\vec{x}_{i}^{g})+
F \times (\vec{x}_{r_{1}}^{g}-\vec{x}_{r_{2}}^{g})
\end{equation}
\item DE/rand-to-best/1
\begin{equation}\label{eqn:derandtobest1}
\vec{v}_{i}^{g}=\vec{x}_{r_{1}}^{g}+F \times (\vec{x}_{best}^{g}-\vec{x}_{r_{1}}^{g})+
F \times (\vec{x}_{r_{2}}^{g}-\vec{x}_{r_{3}}^{g})
\end{equation}
\end{itemize}
where the indices $r_{1}$, $r_{2}$, $r_{3}$, $r_{4}$, and $r_{5}$ are mutually different integers randomly selected from $[1,2,...NP]$  and are also different from $i$, $\vec{x}_{best}^{g}$  denotes the best target vector in the current population, and $F$ refers to the scaling factor.

\emph{Crossover:} After mutation, the crossover operator is performed on each pair of $\vec{x}_{i}^{g}$ and $\vec{v}_{i}^{g}$ to generate a trial vector $\vec{u}_{i}^{g}= [ u_{i,1}^{g},u_{i,2}^{g},...,u_{i,D}^{g}]^{T} $. The binomial crossover is expressed as follows:
\begin{equation}\label{eqn:crossovero}
u_{i,j}^{g}=\left\{
\begin{aligned}
&v_{i,j}^{g},  \text{if} \ rand_{j}\leq CR \text{ or } \ j=j_{rand}\\
&x_{i,j}^{g},  \text{otherwise}
\end{aligned}
\right.
\end{equation}
where $j_{rand}$ is a random integer between $1$ and $D$, $rand_{j}$ is a uniformly distributed random number between 0 and 1 for each $j$, and $CR$ denotes the crossover control parameter. The condition ``$j=j_{rand}$'' makes $\vec{u}_{i}^{g}$ different from $\vec{x}_{i}^{g}$ by at least one dimension.

From Eq.\eqref{eqn:crossovero}, it is easy to see that the crossover operator is implemented dimension by dimension. The updating of the whole trial vector can be described as follows:
\begin{equation}\label{eqn:crossoverw}
\vec{u}_{i}^{g}=\vec{x}_{i}^{g}+\textbf{C}_{r}(\vec{v}_{i}^{g}-\vec{x}_{i}^{g})
\end{equation}
where $\textbf{C}_{r}=diag(s_{1},s_{2},...,s_{D})$, and $ s_{j}=\left\{
\begin{aligned}
&1,  \text{if} \ rand_{j}\leq CR \text{ or } \ j=j_{rand}\\
&0,  \text{otherwise}
\end{aligned}\right.$, $j=1,2,...,D.$

\emph{Selection:} The selection operator is designed to select the better one between $\vec{u}_{i}^{g}$ and $\vec{x}_{i}^{g}$ to enter the next generation. For a minimization problem, it can be described as follows:
\begin{equation}\label{eqn:selection}
\vec{x}_{i}^{g+1}=\left\{
\begin{aligned}
&\vec{u}_{i}^{g}, \text{if}\ f(\vec{u}_{i}^{g})\leq f(\vec{x}_{i}^{g})\\
&\vec{x}_{i}^{g}, \text{otherwise}
\end{aligned}
\right.
\end{equation}

DE has also attracted much attention and a considerable number of DE variants have been proposed. Among them, jDE ~\cite{brest2006self}, SaDE ~\cite{qin2009differential}, and JADE ~\cite{zhang2009jade} are three state-of-the-art DE variants. jDE is a DE with self-adaptive control parameter settings. It encodes the control parameters $F$ and $CR$ into individual level and evolves them. For each individual, the new $F$ is randomly generated within $[0.1,0.9]$ with a probability $\tau_{1}$, and the new $CR$ takes a random value from 0.0 to 1.0 with a probability $\tau_{2}$. SaDE adaptively adjusts the trial vector generation strategies and control parameter settings simultaneously by learning from the previous experience. It maintains a strategy candidate pool which contains four different trial vector generation strategies. Each individual selects a trial vector generation strategy from that pool in an adaptive way to yield its trial vector. JADE is an adaptive DE with an optional external archive. In JADE, the ``DE/current-to-$p$best/1'' mutation operator is proposed which is a generalization of the classical ``DE/current-to-best/1''. This mutation operator exploits the information of multiple best individuals in the population. Moreover, the optional external archive utilizes the difference between the current solutions and the recently explored inferior solutions to produce promising directions. JADE generates $F$ and $CR$ based on their historical record of success.

\section{The Coordinate Systems and Their Related Work}\label{sec:CoSRW}  
\subsection{The coordinate systems}\label{sec:CoS}           
In this subsection, we will introduce the original coordinate system, the Eigen coordinate system, and the difference between them.
\subsubsection{The original coordinate system}\label{sec:OCS}
The original coordinate system is a default coordinate system in most EAs. It is formed by the columns of the unity matrix $\textbf{I}$, and thus is a fixed coordinate system. The evolutionary operators of PSO and DE in Section~\ref{sec:PSO-DE} are conducted in the original coordinate system. By analyzing these evolutionary operators, we find that each of them can be described with the usage of three elements: the coefficients, the diagonal matrixes, and the vectors. Therefore, we propose a new point of view toward how to describe an evolutionary operator in the original coordinate system:
\begin{equation}\label{eqn:commono}
\vec{r}_{O}=\sum_{i=1}^{m}\alpha_{i}\vec{y}_{i}+\sum_{i=1}^{n}\textbf{W}_{i}\vec{z}_{i}
\end{equation}
where $\vec{r}_{O}$ denotes the resultant vector, $m$ and $n$ are nonnegative integers, $\alpha_{i}$  is a coefficient, $\textbf{W}_{i}=diag(w_{1},w_{2},...,w_{D})$, $w_{1}$, $w_{2}$, $...$, $w_{D}$ are real numbers, and $\vec{y}_{i}$ and $\vec{z}_{i}$ mean two vectors in the original coordinate system. Eq.\eqref{eqn:commono} can be considered as a general form of the evolutionary operators in PSO and DE. For example, if $\vec{r}_{O}=\vec{v}_{i}^{g+1}$, $m=1$, $\alpha_{i}=1$, $\vec{y}_{1}=\vec{v}_{i}^{g}$, $n=3$, $\textbf{W}_{1}=c_{1}\textbf{R}_{1}$, $\vec{z}_{1}=\vec{P}_{\text{best}i}^{g}$, $\textbf{W}_{2}=c_{2}\textbf{R}_{2}$, $\vec{z}_{2}=\vec{G}_{\text{best}}^{g}$, $\textbf{W}_{3}=-(c_{1}\textbf{R}_{1}+c_{2}\textbf{R}_{2})$, and $\vec{z}_{3}=\vec{x}_{i}^{g}$, then Eq.\eqref{eqn:commono} is revised to
\begin{equation}\label{eqn:exampleo}
\begin{aligned}
\vec{v}_{i}^{g+1}
&=\vec{v}_{i}^{g}+\left[ c_{1}\textbf{R}_{1}\vec{P}_{\text{best}i}^{g}+ c_{2}\textbf{R}_{2}\vec{G}_{\text{best}}^{g}-( c_{1}\textbf{R}_{1}+c_{2}\textbf{R}_{2})\vec{x}_{i}^{g}\right ]\\
&=\vec{v}_{i}^{g}+c_{1}\textbf{R}_{1}(\vec{P}_{\text{best}i}^{g}-\vec{x}_{i}^{g})+ c_{2}\textbf{R}_{2}(\vec{G}_{\text{best}}^{g}-\vec{x}_{i}^{g})\\
\end{aligned}
\end{equation}
Clearly, Eq.\eqref{eqn:exampleo} is equivalent to Eq.\eqref{eqn:psovw} and both of them are the velocity updating equation in PSO.
Indeed, apart from PSO and DE, Eq.\eqref{eqn:commono} is also an effective way to describe the evolutionary operators in other EA paradigms such as cultural algorithm ~\cite{reynolds1994introduction}, artificial bee colony algorithm ~\cite{karaboga2007powerful}, fireworks algorithm ~\cite{tan2010fireworks}, and brain storm optimization algorithm ~\cite{shi2011brain}.

Note that the right-hand side of Eq.\eqref{eqn:commono} involves two parts: $\sum_{i=1}^{m}\alpha_{i}\vec{y}_{i}$ and $\sum_{i=1}^{n}\textbf{w}_{i}\vec{z}_{i}$. Since the first part is a linear operation of different vectors, it is irrelevant to the coordinate system. In terms of the second part,  $\textbf{W}_{i}$ is a diagonal matrix used for scaling $\vec{z}_{i}$ within the original coordinate system. Since the original coordinate system is a fixed coordinate system, $\textbf{W}_{i}$ can only optimize $\vec{z}_{i}$ in the deterministic directions, thus failing to identify the modality of different function landscapes or even a single function landscape at different optimization stages. As a result, the search process guided by Eq.\eqref{eqn:commono} may not be efficient.\\

\noindent\textbf{Remark 1}: Each evolutionary operator in Section~\ref{sec:PSO-DE} can be rewritten as Eq.\eqref{eqn:commono}. It can be found that the right-hand side of the velocity updating equation in PSO (i.e., Eq.\eqref{eqn:psovw}) and the crossover operator in DE (i.e., Eq.\eqref{eqn:crossoverw}) contains the second part (i.e., $\sum_{i=1}^{n}\textbf{W}_{i}\vec{z}_{i}$), which suggests that these two operators may fail to search efficiently in the original coordinate system.

\subsubsection{The Eigen coordinate system}\label{sec:ECS} 
In this paper, the Eigen coordinate system is established by the columns of an orthogonal matrix $\textbf{B}$, which comes from the Eigen decomposition of the covariance matrix $\textbf{C}$:
\begin{equation}\label{eqn:eigendecomposition}
\textbf{C}=\textbf{BD}^{2}\textbf{B}^{T}
\end{equation}
where $\textbf{B}$ is an orthogonal matrix, $\textbf{B}^{T}$ is the transposed matrix of $\textbf{B}$,  and $\textbf{D}$ is a diagonal matrix. Each column of $\textbf{B}$ is an eigenvector of $\textbf{C}$, and each diagonal element of  $\textbf{D}$ is the square root of an eigenvalue of $\textbf{C}$. The fundamental issue in Eq.\eqref{eqn:eigendecomposition} is how to construct the covariance matrix $\textbf{C}$. In general, the coveriance matrix $\textbf{C}$ is constructed and updated according to the feedback information resulting from the evolution. Therefore, unlike the orginal coordinate system, the Eigen coordinate system is dynamic throughout the evolutionary process, with the aim of suiting the function landscape.

Next, we will discuss how to construct an evolutionary operator in the Eigen coordinate system. It contains three steps. Firstly, $\textbf{B}^{T}$ is applied to transform the vectors in the original coordinate system into the Eigen coordinate system. Subsequently, these vectors in the Eigen coordinate system are combined with the coefficients and diagonal matrixes following Eq.\eqref{eqn:commono}, and thus an offspring vector is obtained. Finally, this offspring vector is transformed back into the original coordinate system by making use of $\textbf{B}$, with the aim of evaluating its fitness. Specifically, an evolutionary operator in the Eigen coordinate system can be described as:
\begin{equation}\label{eqn:commone}
\begin{split}
\vec{r}_{E}
&=\textbf{B}\left (\sum_{i=1}^{m}\alpha_{i}(\textbf{B}^{T}\vec{y}_{i})+\sum_{i=1}^{n}\textbf{W}_{i}(\textbf{B}^{T}\vec{z}_{i})\right )\\
&=\sum_{i=1}^{m}\alpha_{i}\vec{y}_{i}+\sum_{i=1}^{n}\textbf{B}\textbf{W}_{i}\textbf{B}^{T}\vec{z}_{i}
\end{split}
\end{equation}
where $\vec{r}_{E}$ denotes the resultant vector. By comparing Eq.\eqref{eqn:commone} with Eq.\eqref{eqn:commono}, it can be seen that: if we replace $\textbf{W}_{i}$ with  $\textbf{BW}_{i}\textbf{B}^{T}$ on the right-hand side of Eq.\eqref{eqn:commono}, then the evolutionary operator in the original coordinate system is transformed into the corresponding evolutionary operator in the Eigen coordinate system. Compared with $\textbf{W}_{i}\vec{z}_{i}$ , in $\textbf{BW}_{i}\textbf{B}^{T}\vec{z}_{i}$, $\textbf{B}^{T}$  transforms $\vec{z}_{i}$  into the Eigen coordinate system, then $\textbf{W}_{i}$  scales $\textbf{B}^{T}\vec{z}_{i}$ within the Eigen coordinate system, and finally $\textbf{B}$  transforms the vector $\textbf{W}_{i}\textbf{B}^{T}\vec{z}_{i}$ back into the original coordinate system.\\

\noindent\textbf{Remark 2:} PSO's velocity updating equation (i.e., Eq.\eqref{eqn:psovw}) and DE's crossover operator (i.e., Eq.\eqref{eqn:crossoverw}) in the Eigen coordinate system can be expressed as Eq.\eqref{eqn:psovwe} and Eq.\eqref{eqn:crossoverwe}, respectively.
\begin{equation}\label{eqn:psovwe}
\vec{v}_{i}^{g+1}=\vec{v}_{i}^{g}+c_{1}\textbf{BR}_{1}\textbf{B}^{T}(\vec{P}_{\text{best}i}^{g}-
\vec{x}_{i}^{g})+ c_{2}\textbf{BR}_{2}\textbf{B}^{T}(\vec{G}_{\text{best}}^{g}-\vec{x}_{i}^{g})
\end {equation}
\begin{equation}\label{eqn:crossoverwe}
\vec{u}_{i}^{g}=\vec{x}_{i}^{g}+\textbf{BC}_{r}\textbf{B}^{T}(\vec{v}_{i}^{g}-
\vec{x}_{i}^{g})
\end{equation}

\begin{figure}[!t]
    \begin{center}
      \subfigure[]{\includegraphics[width=6cm]{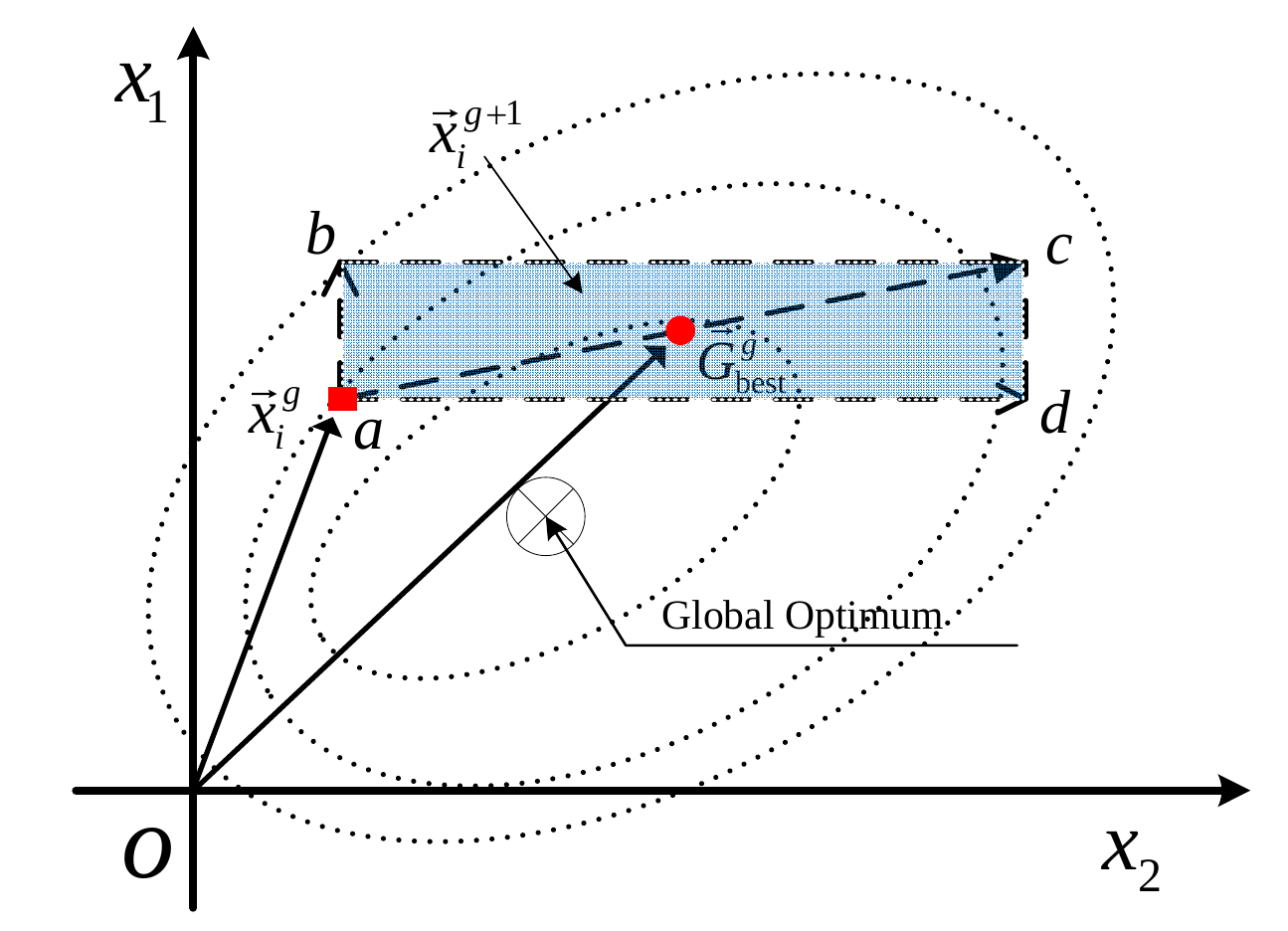}}
        \subfigure[]{\includegraphics[width=6cm]{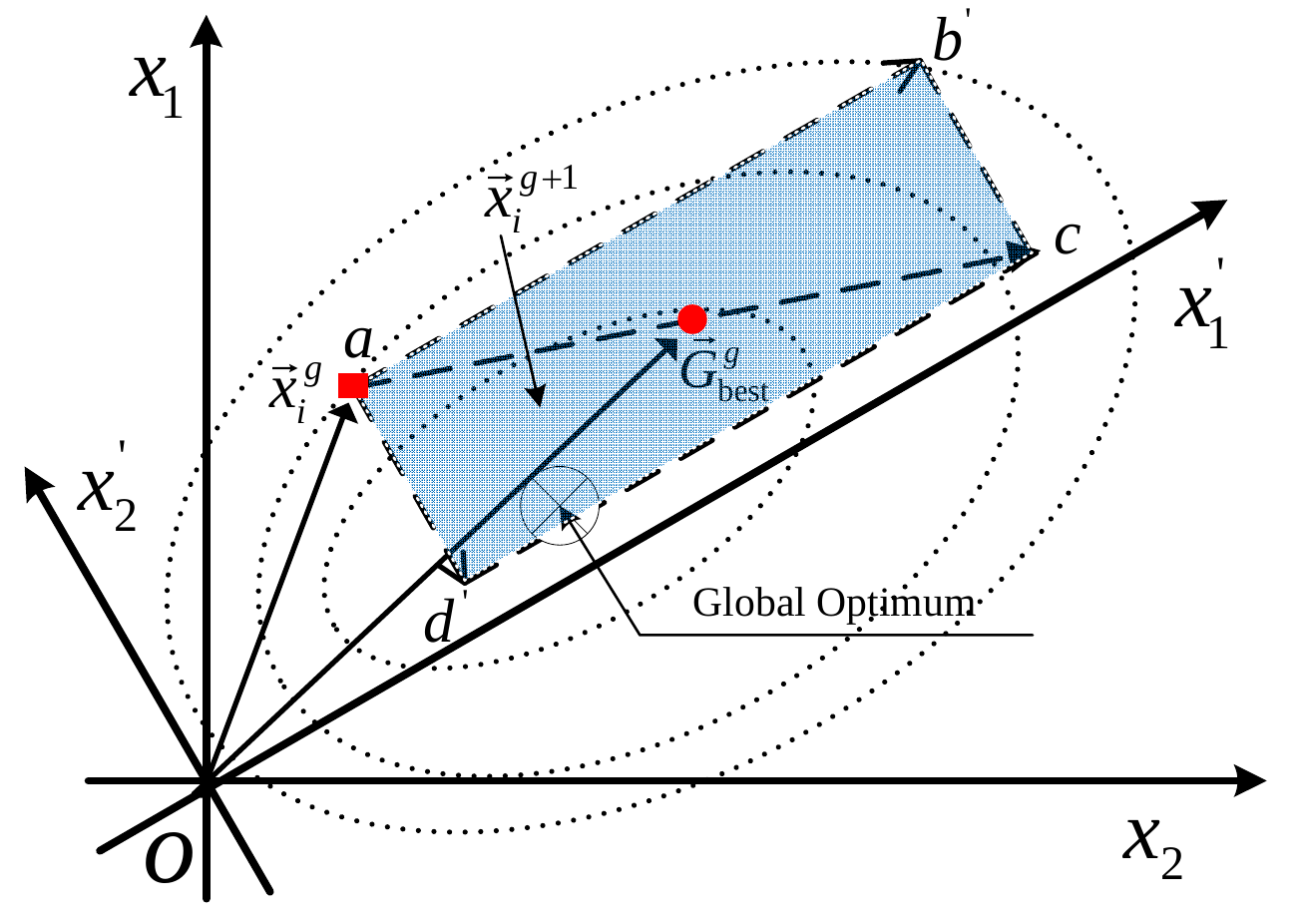}}
       \caption{PSO works in different coordinate systems. In this figure, the dashed ellipses display the contour lines, $\vec{x}_{i}^{g}$ denotes the current position, $\vec{G}_{\text{best}}^{g}$ means the entire swarm's best position, and $\vec{x}_{i}^{g+1}$ is the new position which is located in the blue area. (a) PSO in the original coordinate system (i.e., $ox_{1}x_{2}$). (b) PSO in the Eigen coordinate system (i.e., $ox_{1}^{'}x_{2}^{'}$). }
    \end{center} \label{fig:difference}
\end{figure}

\subsubsection {The difference between the original coordinate system and the Eigen coordinate system}\label{sec:Dif}   
Next, we will investigate EA's search behaviors in the original and Eigen coordinate systems. To make a clear explanation, we take the basic PSO as an example. For simplicity, suppose that the velocity $\vec{v}_{i}^{g}$ of a particle is equal to $\vec{0}$, $c_{1}=c_{2}=2$, the position $\vec{x}_{i}^{g}$ is just its historical best position $\vec{P}_{\text{best}i}^{g}$, and the dimension of the search space is equal to two. As a result, in the original coordinate system, the new velocity $\vec{v}_{i}^{g+1}$ is updated as Eq.\eqref{eqn:copsovo}, and then the new position $\vec{x}_{i}^{g+1}$ is renewed as Eq.\eqref{eqn:copsopo}:
\begin{equation}\label{eqn:copsovo}
\begin{split}
\vec{v}_{i}^{g+1}
&=\vec{v}_{i}^{g}+c_{1}\textbf{R}_{1}(\vec{P}_{\text{best}i}^{g}-\vec{x}_{i}^{g})+c_{2}\textbf{R}_{2} (\vec{G}_{\text{best}}^{g}-\vec{x}_{i}^{g})\\
&=2\textbf{R}_{2}(\vec{G}_{\text{best}}^{g}-\vec{x}_{i}^{g})
\end{split}
\end{equation}
\begin{equation}\label{eqn:copsopo}
\vec{x}_{i}^{g+1}=\vec{x}_{i}^{g}+\vec{v}_{i}^{g+1}
=\vec{x}_{i}^{g}+2\textbf{R}_{2}(\vec{G}_{\text{best}}^{g}-\vec{x}_{i}^{g})
\end{equation}
where $\textbf{R}_{2}=diag(r_{2,1},r_{2,2})$, and $r_{2,1}$ and $r_{2,2}$ are two uniformly distributed random numbers between $0$ and $1$. By replacing $\textbf{R}_{2}$ with $\textbf{BR}_{2}\textbf{B}^{T}$ in Eq.\eqref{eqn:copsopo}, the new position $\vec{x}_{i}^{g+1}$ is generated in the Eigen coordinate system:
\begin{equation}\label{eqn:copsope}
\vec{x}_{i}^{g+1}=\vec{x}_{i}^{g}+2\textbf{BR}_{2}\textbf{B}^{T}(\vec{G}_{\text{best}}^{g}-
\vec{x}_{i}^{g})
\end{equation}
Fig. 1 shows the difference between PSO in the original coordinate system (Fig. 1(a)) and in the Eigen coordinate system (Fig. 1(b)) for an optimization problem with variable correlation. The original coordinate system is fixed and denoted as $ox_{1}x_{2}$.  As pointed out, the Eigen coordinate system is dynamically updated during the evolution. Suppose that for this example the Eigin coordinate system is $ox_{1}^{'}x_{2}^{'}$, which can suit the contour lines well. In Fig. 1, $\vec{x}_{i}^{g+1}$ in the original coordinate system and the Eigen coordinate system is generated as Eq.\eqref{eqn:psoexampleo} and Eq.\eqref{eqn:psoexamplee}, respectively.
\begin{equation}\label{eqn:psoexampleo}
\vec{x}_{i}^{g+1}=\vec{x}_{i}^{g}+2\textbf{R}_{2}(\vec{G}_{\text{best}}^{g}-\vec{x}_{i}^{g})=
\vec{x}_{i}^{g}+r_{2,1}\times\overrightarrow{ab}+r_{2,2}\times\overrightarrow{ad}
\end{equation}
\begin{equation}\label{eqn:psoexamplee}
\vec{x}_{i}^{g+1}=\vec{x}_{i}^{g}+2\textbf{BR}_{2}\textbf{B}^{T}(\vec{G}_{\text{best}}^{g}-
\vec{x}_{i}^{g})=\vec{x}_{i}^{g}+r_{2,1}\times\overrightarrow{ab}^{'}+r_{2,2}\times\overrightarrow{ad}^{'}
\end{equation}
Since $r_{2,1}$ and  $r_{2,2}$ are two uniformly distributed random numbers between $0$ and $1$, $\vec{x}_{i}^{g+1}$ generated in the original and Eigen coordinate systems can be any point in the rectangular areas $abcd$ and $ab^{'}cd^{'}$, respectively. As shown in Fig. 1, $abcd$ does not contain the global optimal solution, while $ab^{'}cd^{'}$ contains the global optimal solution and its neighborhood. This phenomenon signifies that PSO may search more efficiently in the Eigen coordinate system.

\subsection {The related work on the Eigen coordinate system}\label{sec:RW} 
In this paper, the related work on the Eigen coordinate system are classified into two categories, according to the way of conducting the evolutionary operators.

In the first category, the evolutionary operators are  implemented only  in the Eigen coordinate system. In 2001, CMA-ES ~\cite{hansen2001completely} was proposed which samples the offspring population according to:
\begin{equation}\label{eqn:CMA_ESsample}
\begin{aligned}
\vec{x}_{i}^{g+1}
&=\vec{m}^{g}+\sigma^{g}\mathcal{N}(\vec{0},\textbf{\textbf{C}}^{g}),\ i=1,2,...,\lambda\\
&=\vec{m}^{g}+\sigma^{g}(\textbf{C}^{g})^{1/2}\mathcal{N}(\vec{0},\textbf{I}),\ i=1,2,...,\lambda\\
&=\vec{m}^{g}+\sigma^{g}\textbf{B}^{g}\textbf{D}^{g}(\textbf{B}^{g})^{T}\mathcal{N}(\vec{0},\textbf{I}),\ i=1,2,...,\lambda
\end{aligned}
\end{equation}
where $\vec{m}^{g}$ denotes the mean vector of the search distribution at generation $g$, $\sigma^{g}$ denotes the step size, $\textbf{C}^{g}$  refers to a covariance matrix, $\textbf{B}^{g}$ is an orthogonal matrix, $\textbf{D}^{g}$ is a diagonal matrix, $\mathcal{N}(\vec{0},\textbf{C}^{g})$ is a multivariate normal distribution with zero mean and covariance matrix $\textbf{C}^{g}$, and $\mathcal{N}(\vec{0},\textbf{I})$ is a multivariate normal distribution with zero mean and identity covariance matrix $\textbf{I}$. By comparing Eq.\eqref{eqn:CMA_ESsample} with Eq.\eqref{eqn:commone}, it can be found that Eq.\eqref{eqn:CMA_ESsample} is a special case of  Eq.\eqref{eqn:commone}, which means that the sampling operation of CMA-ES only occurs in the Eigen coordinate system. In CMA-ES, this Eigen coordinate system comes from the Eigen decomposition of the covariance matrix $\textbf{C}^{g}$, and two strategies, namely the rank-$\mu$-update strategy and the rank-one-update strategy ~\cite{hansen2016cma}, are designed to adapt $\textbf{C}^{g}$. In the rank-$\mu$-update strategy, a weighted combination of the $\mu$ best out of $\lambda$ offspring is used to compute $\textbf{C}_{\mu}^{g+1}$, which is an estimator of the distribution of the current population:
\begin{equation}\label{eqn:cmarankmu}
\textbf{C}_{\mu}^{g+1}=\sum_{i=1}^{\mu}w_{i}(\vec{x}_{i:\lambda}^{g+1}-\vec{m}^{g})
(\vec{x}_{i:\lambda}^{g+1}-\vec{m}^{g})^{T}\\
\end{equation}
where $w_{i} $ is the $i$th weight coefficient, $\lambda$ is the population size, and $\vec{x}_{i:\lambda}^{(g+1)}$ means the $i$th best individual among the  $\lambda$ offspring. Thereafter, the information from both the previous and current generations are used to compute the covariance matrix $\textbf{C}^{g+1}$:
\begin{equation}\label{eqn:cmarankmuall}
\textbf{C}^{g+1}=(1-c_{\mu})\textbf{C}^{g}+\frac{c_{\mu}}{(\sigma^{g})^{2}}\textbf{C}_{\mu}^{g+1}
\end{equation}
where $c_{\mu}$ is the learning rate for the rank-$\mu$-update strategy. In terms of the rank-one-update strategy, it exploits correlation between consecutive generations and constructs an evolution path to update the covariance matrix. Thus, its implementation is much more complex than the rank-$\mu$-update strategy. These two strategies are combined together in CMA-ES to update the covariance matrix. Since CMA-ES is able to detect the features of the function landscape, it shows a significant superiority over the ordinary ES. To further expand CMA-ES, an adaptive encoding mechanism called AE$_{\textrm{CMA}}$ ~\cite{hansen2009variable} is proposed. In AE$_{\textrm{CMA}}$, a more general approach for covariance matrix adaptation is proposed, which can be applied to ES and estimation of distribution algorithm ~\cite{zhang2008rmeda}. Again, in AE$_{\textrm{CMA}}$, the evolutionary operators are  executed only in the Eigen coordinate system.

In the second category, the evolutionary operators are considered in both the Eigen and original coordinate systems at each generation of EAs. For instance, DE/eig ~\cite{guo2015enhancing} and CoBiDE ~\cite{wang2014differential} implement the crossover operator of DE in both the Eigen and original coordinate systems in a random manner.
As a result, similar to the classical DE, one trial vector is created for one target vector. In DE/eig, all individuals from the current generation are used to compute the covariance matrix:
\begin{equation}\label{eqn:deeig}
\textbf{C}^{g+1}=\frac{1}{N\!P\!-\!1}\!\sum_{i=1}^{N\!P}\!(\vec{x}_{i}^{g}\!-\!\frac{1}{N\!P}\!\sum_{j=1}^{NP}\!\vec{x}_{j}^{g}
)(\vec{x}_{i}^{g}\!-\!\frac{1}{N\!P}\!\sum_{j=1}^{N\!P}\!\vec{x}_{j}^{g})^{T}
\end{equation}
where $NP$ is the population size, and $\vec{x}_{i}^{g}$ and $\vec{x}_{j}^{g}$ mean the $i$th and $j$th individuals, respectively. While in CoBiDE, the $NP^{'}$ best out of the individuals from the current population are employed to update the covariance matrix:
\begin{equation}\label{eqn:cobide}
\begin{split}
\textbf{C}^{g+1}
&=\frac{1}{N\!P^{'}\!-\!1}\sum_{i=1}^{N\!P^{'}}
(\vec{x}_{i:N\!P}^{g}\!-\!\frac{1}{N\!P^{'}}\sum_{j=1}^{N\!P^{'}}\vec{x}_{j:N\!P}^{g})\times\\
&(\vec{x}_{i:N\!P}^{g}\!-\!\frac{1}{N\!P^{'}}\sum_{j=1}^{N\!P^{'}}\vec{x}_{j:N\!P}^{g})^{T}
\end{split}
\end{equation}
where $NP^{'}=ps\times NP$, $ps\in[0,1]$ is a user-defined parameter, and $\vec{x}_{i:NP}^{g}$ and $\vec{x}_{j:NP}^{g}$ denote the $i$th and $j$th best individuals, respectively. From Eq.\eqref{eqn:deeig} and Eq.\eqref{eqn:cobide}, it can be seen that only the current population distribution information is utilized to compute the covariance matrix. Very recently, a novel DE framework called CPI-DE ~\cite{wang2016utilizing} is proposed. In CPI-DE, DE's crossover operator is executed in both the Eigen and original coordinate systems in a deterministic manner and, consequently, two trial vectors are generated for each target vector. Thereafter, the best one among the target vector and its two trial vectors will survive into the next generation. The covariance matrix in CPI-DE is estimated by the rank-$NP$-update strategy, which can be regarded as an extension of the rank-$\mu$-update strategy in CMA-ES. This rank-$NP$-update strategy contains two steps. In the first step, the $NP$ best out of $2 \times NP$ offspring (note that in CPI-DE, the offspring population consists of $2 \times NP$ trial vectors) are used to estimate the current population distribution:
\begin{equation}\label{eqn:cpide1}
\textbf{C}_{NP}^{g+1}=\sum_{i=1}^{NP}w_{i}(\vec{x}_{i:2*NP}^{g+1}-\vec{m}^{g})
(\vec{x}_{i:2*NP}^{g+1}-\vec{m}^{g})^{T}
\end{equation}
where $w_{i}$ is the $i$th weight coefficient and $\vec{x}_{i:2*NP}^{g+1}$  represents the $i$th best individual in the offspring population. In the second step, the population distribution information from the current and historical generations are used to adapt the covariance matrix:
\begin{equation}\label{eqn:cpide2}
\textbf{C}^{g+1}=(1-c_{NP})\textbf{C}^{g}+\frac{c_{\mu}}{(\sigma^{g})^{2}}\textbf{C}_{NP}^{g+1}
\end{equation}
where $c_{NP}$ is the learning rate and $\sigma^{g}$ is the step size. It is claimed in CPI-DE ~\cite{wang2016utilizing} that there
is no necessary to adapt the step size for DE, since DE has a different search pattern with ES. In fact, $\sigma^{g}$ is set to 1 in CPI-DE, which means that the covariance matrix is of equal importance at each generation. It is necessary to note that CPI-DE does not utilize the  rank-one-update strategy. The reason is that the rank-one-update strategy plays a less important role when the population size is large, and DE usually maintains a relatively large population compared with ES. Besides, the rank-one-update strategy is much more complex than the rank-$\mu$-update strategy. Therefore, by eliminating the rank-one-update strategy, the adaptation of the covariance matrix in CPI-DE becomes simpler. There is an agreement from the above three attempts: the usage of both the Eigen and original coordinate systems at each generation can reach better performance than the usage of one of them during the whole evolution.

Our work in this paper falls into the second category. Moreover, the Eigen and original coordinate systems are adaptively tuned as the evolution proceeds.

\section{Proposed Approach}\label{sec:PA}    
\subsection {Motivation and general framework}   
We continue the work on the coordinate systems and propose a novel framework named ACoS. The motivation of ACoS comes from three aspects:
\begin{itemize}
  \item
A large population can provide more information to estimate the Eigen coordinate system, compared with a small population. However, given the maximum number of fitness evaluations, the increase of the population size will lead to the decrease of the generation number, which might cause incomplete convergence of EAs. Consequently, it is necessary to design a mechanism to strike the balance between the accuracy of estimation and the convergence performance.
  \item
As introduced in Section~\ref{sec:RW}, some researchers have recognized the importance of combining the original coordinate system with the Eigen coordinate system in the evolutionary computation research community. However, the current methods adjust these two coordinate systems in either a random way or a deterministic way. How to exploit the feedback information from the evolution to adaptively tune them has not yet been investigated.
  \item
The coordinate systems play a very important role in the performance of EAs. Note, however, that in current studies the coordinate systems have been applied to enhance the performance of few EA paradigms (e.g., ES and DE). It is an interesting topic to boost the research on the coordinate systems to other EA paradigms.
\end{itemize}

ACoS aims at addressing the above three issues. In ACoS, an additional archiving mechanism is designed to maintain the offspring not only in the current generation but also in the past several generations. Therefore, sufficient information can be obtained to estimate an appropriate Eigen coordinate system without adding the population size and reducing the generation number. As a result, ACoS achieves a balance between the accuracy of estimation and the convergence performance. Afterward, the Eigen and original coordinate systems are selected in an adaptive way (rather than a random or deterministic way) according to a probability vector, which is updated based on the collected information from the offspring. ACoS can be readily applied to various EAs, and in this paper we consider two of the most popular EA paradigms: PSO and DE. The general framework of ACoS has been given in \textbf{Algorithm 1}.

\begin{algorithm}[!t]   
   \caption{The framework of ACoS}\label{alg:ACoS}
   \begin{algorithmic}[1]
   \STATE Initialize $g=0$, $\textbf{P}^{0}=\{\vec{x}_{1}^{0},\vec{x}_{2}^{0},...,\vec{x}_{NP}^{0 } \}$,  archive  $\textbf{A}=\emptyset$, and $\textbf{C}^{0}=\textbf{B}^{0}=\textbf{I}$;
   \STATE Initialize the probability vector$\vec{p}=(p_{1},p_{2},...,p_{NP})=(0.5,0.5,...,0.5)$;\\
   \WHILE {the termination criterion is not met}
   \FOR {$i=1$ to $NP$}
   \IF {$rand\leq p_{i}$}
   \STATE Implement the mutation and crossover operators of EAs in the Eigen coordinate system to generate the $i$th offspring;
   \ELSE
   \STATE Implement the mutation and crossover operators of EAs in the original coordinate system to generate the $i$th offspring;
   \ENDIF
   \ENDFOR
   \STATE Evaluate the offspring population;
   \STATE Implement the selection operator of EAs to get $\textbf{P}^{g+1}$;
   \STATE Update $\textbf{A}$, $\textbf{C}^{g+1}$, and $\textbf{B}^{g+1}$ based on Section~\ref{sec:UECS};
   \STATE Update $\vec{p}$ according to Section~\ref{sec:UPV};
   \STATE $g = g +1 $;
   \ENDWHILE
\end{algorithmic}
\end{algorithm}

In \textbf{Algorithm 1}, $rand$ denotes a uniformly distributed random number on the interval $[0,1]$. In the initialization process, the population $\textbf{P}^{0}=\{\vec{x}_{1}^{0}, \vec{x}_{2}^{0},...,\vec{x}_{NP}^{0}\}$ is randomly sampled from the search space, the archive $\textbf{A}$ is initialized to be empty, the covariance matrix $\textbf{C}^{0}$ and the orthogonal matrix $\textbf{B}^{0}$ are set to be the unity matrix $\textbf{I}$, and the probability vector $\vec{p}=(p_{1},p_{2},...,p_{NP})$ is initialized to be $\vec{p}=(0.5,0.5,...,0.5)$. During the evolution, to generate the $i$th offspring, the mutation and crossover operators of EAs are implemented in the Eigen and original coordinate systems with the probabilities $p_{i}$ and $1-p_{i}$, respectively. Afterward, the offspring population is evaluated and the selection operator of EAs is executed to obtain $\textbf{P}^{g+1}$. Subsequently, $\textbf{A}$, $\textbf{C}^{g+1}$, and  $\textbf{B}^{g+1}$ are updated according to Section~\ref{sec:UECS}. Finally, $\vec{p}$ is renewed according to Section~\ref{sec:UPV}.

Obviously, ACoS is different from the canonical EAs due to the simultaneous use and adaptive tuning of the Eigen and original coordinate systems. Next, we will introduce two core components of ACoS: the updating of the Eigen coordinate system and the updating of $\vec{p}$.


\subsection{The updating of the Eigen coordinate system}\label{sec:UECS}  
The Eigen coordinate system is updated by making use of an additional archiving mechanism and the rank-$\mu$-update strategy ~\cite{hansen2016cma}.

The additional archiving mechanism adopts an external archive $\textbf{A}$ to store the offspring in both the current generation and the past several generations. It is because the search area may not change dramatically in the continuous several generations of EAs, and thus the offspring in the past several generations, other than the offspring in the current generation, can also provide the important information to estimate an appropriate Eigen coordinate system. Actually, the implementation of this additional archiving mechanism is very simple. Firstly, $\textbf{A}$ is initialized to be an empty set. Then at each generation, the newly generated offspring are added into $\textbf{A}$. If the archive size (called $AS$) exceeds a certain threshold, say $3\times NP$, then the earlier offspring in  $\textbf{A}$ will be removed based on the ``first-in-first-out'' rule to keep the archive size at $3\times NP$. Note that unlike the main population $\textbf{P}^{g}$, $\textbf{A}$ does not undergo any evolutionary operators. Therefore, this additional archiving mechanism can obtain sufficient information to estimate the Eigen coordinate system while never affecting the population size and the generation number.



Subsequently, the rank-$\mu$-update strategy extracts the population distribution information from $\textbf{A}$. The previous research has demonstrated that the rank-$\mu$-update strategy is an efficient technique to adapt the covariance matrix ~\cite{hansen2016cma}. In this paper, the size of $\textbf{A}$ is larger than that of $\textbf{P}^{g}$. Therefore, the rank-$\mu$-update can benefit from this relatively larger size to get a reliable estimator of the covariance matrix. Before executing the rank-$\mu$-update strategy, we need to initialize the mean vector of the search distribution $\vec{m}^{g}$ in Eq.\eqref{eqn:cmarankmu} and the covariance matrix $\textbf{C}^{g}$. In this paper, $\vec{m}^{0}$ is set to be a randomly generated point in the search space and $\textbf{C}^{0}$ is set to be the unity matrix $\textbf{I}$. Then, at generation $g+1$, $\vec{m}^{g+1}$ is updated according to:
\begin{equation}\label{eqn:acosmean}
\vec{m}^{g+1}=\sum_{i=1}^{\mu}\omega_{i}\vec{a}_{i:AS}
\end{equation}
where $\mu=AS/2$ is the number of the selected solutions, $\vec{a}_{i:AS}$ denotes the $i$th best solution out of $\textbf{A}$ (i.e., $f(\vec{a}_{1:AS})\leq f(\vec{a}_{2:AS})\leq...\leq f(\vec{a}_{\mu:AS})$), and $\omega_{i}$ refers to the $i$th weight coefficient computed as:
\begin{equation}\label{eqn:acosomega}
\omega_{i}=\frac{\ln (\mu+0.5)-\ln i}{n\ln(\mu+0.5)-\sum_{j=1}^{\mu}\ln i}, i=1,2,...,\mu
\end{equation}
Afterward, an estimator of the current population distribution  $\textbf{C}_{\mu}^{g+1}$ is obtained by:
\begin{equation}\label{eqn:acoscmu}
\textbf{C}_{\mu }^{g+1}=\sum_{i=1}^{\mu}\omega_{i}(\vec{a}_{i:AS}-\vec{m}^{g})
(\vec{a}_{i:AS}-\vec{m}^{g})^{T}
\end{equation}
Finally, the covariance matrix $\textbf{C}^{g+1}$ is updated by making use of the cumulative population distribution information:
\begin{equation}\label{eqn:acosc}
\textbf{C}^{g+1}=(1-c_{\mu})\textbf{C}^{g}+c_{\mu}\textbf{C}_{\mu }^{g+1}
\end{equation}
where $c_{\mu}\approx\frac{1}{3}\mu_{eff}/D^2$ denotes the learning rate, $\mu_{eff}=(\sum_{i=1}^{\mu}\omega_{i}^{2})^{-1}$  is the variance effective selection mass, and $D$ is the dimension of the search space.

After $\textbf{C}^{g+1}$ is obtained, an Eigen decomposition is performed on $\textbf{C}^{g+1}$ according to Eq.\eqref{eqn:eigendecomposition} to produce the orthogonal matrix $\textbf{B}^{g+1}$, the columns of which form the Eigen coordinate system.

\subsection {The updating of the probability vector}\label{sec:UPV} 
The probability vector $\vec{p}=(p_{1},p_{2},...,p_{NP})$  determines the selection ratio of each coordinate system for each individual.
Since there is no priori knowledge about the characteristics of the function landscapes, the Eigen and original coordinate systems are considered to be of equal importance at the beginning of evolution, i.e., $\vec{p}=(0.5,0.5,...,0.5)$. Then, $\vec{p}$ is adaptively updated during the evolution according to the collected information derived from the offspring.

In this paper, we collect the information including which coordinate system is used to generate the offspring and how about the quality of the generated offspring. It is easy to identify which coordinate system is used to produce the offspring. However, how to measure the quality of the offspring is usually dependent on a specific EA. For PSO, if a particle's new position is better than its personal historical best position, then the offspring performs better, otherwise, it  performs worse. In terms of DE, if the trial vector outperforms its corresponding target vector, then the offspring performs better; otherwise, it performs worse. Without loss of generality, the collected information derived from the offspring can be categorized into four cases:
 \begin{itemize}
   \item
   \emph{Eigen coordinate system is better}: the Eigen coordinate system is used to generate the offspring and the offspring performs better;
   \item
   \emph{Eigen coordinate system is worse}: the Eigen coordinate system is used to generate the offspring but the offspring performs worse;
   \item
   \emph{Original coordinate system is better}: the original coordinate system is used to generate the offspring and the offspring performs better;
   \item
   \emph{Original coordinate system is worse}: the original coordinate system is used to generate the offspring  but the offspring performs worse.
 \end{itemize}

\begin{algorithm}[!t]  
    \caption{The updating of the probability vector}\label{alg:vector}
    \begin{algorithmic}[1]
    \newcommand{\SWITCH}[1]{\STATE \textbf{switch} (#1)}
    \newcommand{\ENDSWITCH}{\STATE \textbf{end switch}}
    \newcommand{\CASE}[1]{\STATE \textbf{case} #1\textbf{:} \begin{ALC@g}}
    \newcommand{\ENDCASE}{\end{ALC@g}}
    \newcommand{\CASELINE}[1]{\STATE \textbf{case} #1\textbf{:} }
    \newcommand{\DEFAULT}{\STATE \textbf{default:} \begin{ALC@g}}
    \newcommand{\ENDDEFAULT}{\end{ALC@g}}
    \newcommand{\DEFAULTLINE}[1]{\STATE \textbf{default:} }
   \SWITCH {the case of the collected information from the offspring}
   \CASE {\emph{Eigen coordinate system is better}}
   \STATE $p_{i}\leftarrow p_{i}+r(p_{i})$;
   \ENDCASE
   \CASE {\emph{Eigen coordinate system is worse}}
   \STATE $p_{i}\leftarrow p_{i}-\eta \times r(p_{i})$;
   \ENDCASE
   \CASE {\emph{original coordinate system is better}}
   \STATE $p_{i}\leftarrow p_{i}-r(1-p_{i})$;
   \ENDCASE
   \CASE {\emph{original coordinate system is worse}}
   \STATE $p_{i}\leftarrow p_{i}+\eta \times r(1-p_{i})$;
   \ENDCASE
   \ENDSWITCH
\end{algorithmic}
\end{algorithm}

These four cases have been considered fully in \textbf{Algorithm 2} to adaptively update $\vec{p}$. The main principle behind \textbf{Algorithm 2} is the ``use it or lose it'' rule: if one coordinate system is used to generate the offspring and the offspring performs better, the selection ratio for this coordinate system will increase; otherwise, the selection ratio for this coordinate system will decrease. More specifically, for the $i$th individual:
\begin{itemize}
  \item  In the case of \emph{Eigen coordinate system is better}, a reward $r(p_{i})$ is added into $p_{i}$. $r(\bullet)$ denotes a reward function defined as $r(x)=\varepsilon(1-x)e^{-2x}, \ x\in [0,1]$. In this reward function, $\varepsilon$ is a constriction factor to clamp the reward value into $[0,\varepsilon]$, and $(1-x)e^{-2x} $ is a concave function whose value decreases from 1 to 0 when the variable $x$ increases from 0 to 1. As a result, a larger $p_{i}$ will receive a smaller $r(p_{i})$. It is reasonable since a larger  $p_{i}$  means that the Eigen coordinate system already has more potential to be chosen, and a smaller reward would restrain the dramatic increasing of $p_{i}$ and adapt $p_{i}$ to a proper value in a more robust way.
  \item In the case of \emph{Eigen coordinate system is worse}, a punishment $\eta \times r(p_{i})$ is added into $p_{i}$.
        In $\eta \times r(p_{i})$, $\eta$ is a punishment coefficient on the interval $(0,1)$. Therefore, $\eta \times r(p_{i})$ is smaller than $r(p_{i})$, which implies that the case \emph{Eigen coordinate system is worse} has less influence on $p_{i}$ than the case \emph{Eigen coordinate system is better} at one time. The reason is the following. EA is a trial-and-error method and the case \emph{Eigen coordinate system is worse} is more likely to happen compared with the case \emph{Eigen coordinate system is better}. Therefore, the more likely occurred case (i.e., \emph{Eigen coordinate system is worse}) should have less influence on $p_{i}$ than the less likely occurred case (i.e., \emph{Eigen coordinate system is better}) at one time due to the fact that these two cases' whole effects on $p_{i}$ should be similar.
  \item In the case of \emph{original coordinate system is better}, the selection ratio of the original coordinate system will increase and, therefore, $p_{i}$  will decrease. The reduced value is equal to $r(1-p_{i})$.
  \item In the case of \emph{original coordinate system is worse}, the selection ratio of the original coordinate system will decrease and  $p_{i}$ thus will increase. The increased value is equal to $\eta \times r(1-p_{i})$.
\end{itemize}

\subsection {The application of ACoS in PSO and DE}\label{sec:Extension}    
ACoS has a simple structure and can be easily applied to various EAs. For a specific EA, if it is under the framework of ACoS, it will dynamically select one of the Eigen and original coordinate systems according to $\vec{p}$ to generate the offspring. Since the updating of the Eigen coordinate system and $\vec{p}$ has been introduced previously, when implementing a specific EA under the framework of ACoS, we only need to consider how to generate the offspring in different coordinate systems and how to use the selection operator. In this paper, we apply ACoS to two of the most popular EA paradigms, namely PSO and DE.

For PSO, the offspring are generated via the velocity updating equation and the position updating equation. These two equations in the original coordinate system have been given in Eq.\eqref{eqn:psovw} and Eq.\eqref{eqn:psopw}, respectively. According to Section~\ref{sec:CoSRW}, Eq.\eqref{eqn:psopw} is irrelevant to the coordinate systems. With respect to Eq.\eqref{eqn:psovw}, it depends on the coordinate systems and its implementation in the Eigen coordinate system has been given in Eq.\eqref{eqn:psovwe}. It is necessary to note that PSO does not employ the selection operator and, therefore, the selection operator in Step (12) of \textbf{Algorithm 1} can be eliminated.

For DE, the offspring are produced through the mutation and crossover operators. These two operators in the original coordinate system have been given in Eqs.\eqref{eqn:derand1}-\eqref{eqn:derandtobest1} and Eq.\eqref{eqn:psopw}, respectively. In fact, the mutation operator is independent of the coordinate systems, while the crossover operator relies on the coordinate systems, the implementation of which in the Eigen coordinate system has been given in Eq.\eqref{eqn:crossoverwe}. In addition, the selection operator of ACoS is the same with that of the original DE.

\subsection {Characteristics of ACoS}\label{sec:Character}   
Next, we compare ACoS with other related work introduced in Section~\ref{sec:CoSRW}. Compared with CMA-ES which samples all the individuals in the Eigen coordinate system, ACoS has some advantages listed as follows:
\begin{itemize}
  \item It makes use of both the Eigen and original coordinate systems. The Eigen coordinate system enables EAs to identify the modality of the fitness landscape and enhance the search efficiency, while the original coordinate system can maintain the superiority of the original EAs.
  \item The updating of the Eigen coordinate system in ACoS is simpler. ACoS eliminates the much more complex rank-one-update strategy and only adopts the rank-$\mu$-update strategy to estimate the Eigen coordinate system. In addition, an additional archiving mechanism with negligible computational cost is designed to improve the estimation accuracy.
  \item ACoS can be readily applied to other EAs. This can be attributed to the fact that the step-size control, which plays a very important role in CMA-ES, can be ignored in many other EAs due to their different search patterns with CMA-ES.
\end{itemize}

Compared with DE/eig, CoBiDE and CPI-DE which focus on enhancing DE's performance, ACoS has the following advantages:
\begin{itemize}
\item ACoS is designed to improve the performance of not only DE but also other EAs.
  \item To update the Eigen coordinate system, DE/eig and CoBiDE only utilize the current population distribution information, therefore the established Eigen coordinate system might be inappropriate due to insufficient information. In CPI-DE and ACoS, the cumulative population distribution information is used to update the Eigen coordinate system. Note, however, that ACoS employs an additional archiving mechanism which can obtain more sufficient information while having no influence on the population size and the generation number.
  \item Although both the Eigen and original coordinate systems are utilized in DE/eig, CoBiDE, CPI-DE, and ACoS, DE/eig and CoBiDE adjust these two coordinate systems in a random manner which ignores the feedback information from the evolutionary search and, therefore, is not well suited for different kinds of fitness landscapes. In addition, CPI-DE generates two offspring for each target vector, one in the Eigen coordinate system and the other in the original coordinate system, which inevitably spends more fitness evaluations at each generation. In contrast, ACoS adapts these two coordinate systems in an adaptive way as the evolution proceeds, thus exploiting the feedback information and producing only one offspring for each target vector simultaneously.

\end{itemize}

\begin{table*}[!t]
\scriptsize
\centering
\addtolength{\tabcolsep}{-4pt}
\renewcommand{\arraystretch}{1.25}
\caption{Experimental results of PSO-w, ACoS-PSO-w, PSO-cf, and ACoS-PSO-cf over 51 independent runs on 30 test functions with 30D from IEEE CEC2014 using 300,000 FEs.}\label{tbl:pso30D}
\newcommand{\minitab}[2][l]{\begin{tabular}{#1}#2\end{tabular}}
\newcommand{\tabincell}[2]{\begin{tabular}{@{}#1@{}}#2\end{tabular}}
\begin{tabular}[width=\linewidth]{r|c|c|c||c|c}
\hline
\hline
\multicolumn{2}{c|}{\multirow{2}{*}{\tabincell{c}{Test Functions with 30D\\ from IEEE CEC2014}}} & PSO-w & ACoS-PSO-w & PSO-cf & ACoS-PSO-cf \\
\multicolumn{2}{c|}{} & Mean Error$\pm$Std Dev & Mean Error$\pm$Std Dev & Mean Error$\pm$Std Dev & Mean Error$\pm$Std Dev \\
\hline
\multicolumn{1}{c|}{\multirow{3}{*}{\tabincell{c}{Unimodal\\Functions}}} & $cf_{1}$ & 1.53E+08$\pm$1.34E+08$-$ & 1.86E+06$\pm$3.60E+06 & 6.68E+07$\pm$7.48E+07$-$ & 1.08E+03$\pm$2.62E+03 \\
\cline{2-6}
\multicolumn{1}{c|}{} & $cf_{2}$  & 1.67E+10$\pm$7.67E+09$-$ & 1.00E+03$\pm$5.14E+03 & \textbf{7.77E+09$\pm$5.94E+09$-$} & \textbf{0.00E+00$\pm$0.00E+00} \\
\cline{2-6}
\multicolumn{1}{c|}{} & $cf_{3}$  & 4.72E+04$\pm$3.37E+04$-$ & 1.75E+01$\pm$1.23E+02 & \textbf{1.17E+04$\pm$1.59E+04$-$} & \textbf{0.00E+00$\pm$0.00E+00} \\
\cline{1-6}
\multicolumn{1}{c|}{\multirow{13}{*}{\tabincell{c}{Simple \\ Multimodal \\ Functions}}} & $cf_{4}$ & 1.28E+03$\pm$9.00E+02$-$ & 1.00E+02$\pm$6.30E+01 & 8.37E+02$\pm$9.65E+02$-$ & 1.83E+01$\pm$2.88E+01\\
\cline{2-6}
\multicolumn{1}{c|}{} & $cf_{5}$  & 2.07E+01$\pm$1.35E-01$\approx$ & 2.07E+01$\pm$1.22E-01 & 2.02E+01$\pm$2.83E-01$+$ & 2.07E+01$\pm$1.41E-01 \\
\cline{2-6}
\multicolumn{1}{c|}{} & $cf_{6}$  & 2.07E+01$\pm$2.92E+00$-$ & 1.97E+01$\pm$3.30E+00 & 2.10E+01$\pm$3.35E+00$-$ & 1.81E+01$\pm$4.17E+00 \\
\cline{2-6}
\multicolumn{1}{c|}{} & $cf_{7}$  & 1.76E+02$\pm$7.89E+01$-$ & 1.06E+01$\pm$7.76E+00 & 1.05E+02$\pm$7.63E+01$-$ & 1.11E-02$\pm$1.34E-02 \\
\cline{2-6}
\multicolumn{1}{c|}{} & $cf_{8}$ & 9.63E+01$\pm$2.67E+01$-$ & 7.19E+01$\pm$1.82E+01 & 9.59E+01$\pm$3.56E+01$-$ & 8.62E+01$\pm$2.35E+01 \\
\cline{2-6}
\multicolumn{1}{c|}{} & $cf_{9}$ & 1.50E+02$\pm$3.10E+01$-$ & 1.19E+02$\pm$2.38E+01 & 1.38E+02$\pm$4.02E+01$-$ & 9.55E+01$\pm$2.65E+01 \\
\cline{2-6}
\multicolumn{1}{c|}{} & $cf_{10}$ & 3.36E+03$\pm$7.42E+02$-$ & 2.69E+03$\pm$5.82E+02 & 2.72E+03$\pm$8.03E+02$+$ & 3.04E+03$\pm$6.17E+02 \\
\cline{2-6}
\multicolumn{1}{c|}{} & $cf_{11}$ & 3.65E+03$\pm$7.20E+02$-$ & 3.45E+03$\pm$7.27E+02 & 3.62E+03$\pm$6.73E+02$\approx$ & 3.63E+03$\pm$5.65E+02 \\
\cline{2-6}
\multicolumn{1}{c|}{} & $cf_{12}$ & 7.19E-01$\pm$5.24E-01$+$ & 1.17E+00$\pm$6.17E-01 & 3.26E-01$\pm$1.08E-01$+$ & 6.81E-01$\pm$4.43E-01 \\
\cline{2-6}
\multicolumn{1}{c|}{} & $cf_{13}$ & 3.03E+00$\pm$1.25E+00$-$ & 6.99E-01$\pm$9.79E-02 & 2.39E+00$\pm$1.25E+00$-$ & 4.32E-01$\pm$1.04E-01 \\
\cline{2-6}
\multicolumn{1}{c|}{} & $cf_{14}$ & 5.11E+01$\pm$2.68E+01$-$ & 1.19E+00$\pm$2.59E-01 & 4.19E+01$\pm$3.13E+01$-$ & 5.77E-01$\pm$2.56E-01 \\
\cline{2-6}
\multicolumn{1}{c|}{} & $cf_{15}$ & 7.52E+04$\pm$2.33E+05$-$ & 1.08E+03$\pm$3.81E+03 & 1.03E+04$\pm$3.49E+04$-$ & 5.24E+00$\pm$1.70E+00 \\
\cline{2-6}
\multicolumn{1}{c|}{} & $cf_{16}$ & 1.13E+01$\pm$6.30E-01$\approx$ & 1.15E+01$\pm$5.00E-01 & 1.14E+01$\pm$6.21E-01$\approx$ & 1.13E+01$\pm$6.22E-01 \\
\cline{1-6}
\multicolumn{1}{c|}{\multirow{6}{*}{\tabincell{c}{Hybrid \\ Functions}} }& $cf_{17}$ & 4.99E+06$\pm$5.88E+06$-$ & 8.00E+04$\pm$2.34E+05 & 2.26E+06$\pm$4.46E+06$-$ & 1.73E+03$\pm$3.74E+02 \\ \cline{2-6}
\multicolumn{1}{c|}{} & $cf_{18}$ & 2.16E+08$\pm$4.65E+08$-$ & 5.48E+03$\pm$5.47E+03 & 7.51E+07$\pm$2.94E+08$-$ & 9.38E+03$\pm$8.31E+03 \\
\cline{2-6}
\multicolumn{1}{c|}{} & $cf_{19}$ & 6.84E+01$\pm$6.64E+01$-$ & 2.60E+01$\pm$2.66E+01 & 6.00E+01$\pm$5.33E+01$-$ & 1.13E+01$\pm$2.07E+00 \\
\cline{2-6}
\multicolumn{1}{c|}{} & $cf_{20}$ & 1.87E+04$\pm$2.46E+04$-$ & 5.62E+02$\pm$8.51E+02 & 6.61E+03$\pm$1.58E+04$-$ & 3.21E+02$\pm$1.33E+02 \\
\cline{2-6}
\multicolumn{1}{c|}{} & $cf_{21}$ & 8.45E+05$\pm$1.06E+06$-$ & 2.56E+04$\pm$9.06E+04 & 8.93E+05$\pm$3.76E+06$-$ & 1.20E+03$\pm$7.46E+02 \\
\cline{2-6}
\multicolumn{1}{c|}{} & $cf_{22}$ & 6.53E+02$\pm$3.21E+02$-$ & 4.72E+02$\pm$1.97E+02 & 6.36E+02$\pm$2.39E+02$-$ & 4.93E+02$\pm$1.97E+02 \\
 \cline{1-6}
\multicolumn{1}{c|}{\multirow{8}{*}{\tabincell{c}{Composition \\ Functions }}} & $cf_{23}$ & 4.03E+02$\pm$6.51E+01$-$ & 3.18E+02$\pm$7.06E+00 & 3.65E+02$\pm$4.81E+01$-$ & 3.15E+02$\pm$4.54E-13 \\
 \cline{2-6}
\multicolumn{1}{c|}{} & $cf_{24}$ & 2.70E+02$\pm$2.30E+01$-$ & 2.45E+02$\pm$7.89E+00 & 2.61E+02$\pm$2.35E+01$-$ & 2.42E+02$\pm$7.12E+00 \\
\cline{2-6}
\multicolumn{1}{c|}{} & $cf_{25}$ & 2.19E+02$\pm$1.05E+00$-$ & 2.05E+02$\pm$2.49E+00 & 2.16E+02$\pm$8.63E+00$-$ & 2.07E+02$\pm$5.45E+00 \\
\cline{2-6}
\multicolumn{1}{c|}{} & $cf_{26}$ & 1.30E+02$\pm$6.78E+01$-$ & 1.20E+02$\pm$6.50E+01 & 1.32E+02$\pm$6.43E+01$-$ & 1.14E+02$\pm$5.36E+01 \\
\cline{2-6}
\multicolumn{1}{c|}{} & $cf_{27}$ & 1.05E+03$\pm$1.81E+02$-$ & 9.18E+02$\pm$2.17E+02 & 9.16E+02$\pm$2.77E+02$\approx$ & 9.10E+02$\pm$2.07E+02 \\
\cline{2-6}
\multicolumn{1}{c|}{} & $cf_{28}$ & 1.93E+03$\pm$4.66E+02$-$ & 1.34E+03$\pm$2.97E+02 & 1.89E+03$\pm$4.67E+02$-$ & 1.65E+03$\pm$4.43E+02 \\
\cline{2-6}
\multicolumn{1}{c|}{} & $cf_{29}$ & 1.70E+07$\pm$1.54E+07$-$ & 8.49E+06$\pm$1.05E+07 & 1.64E+07$\pm$1.21E+07$-$ & 8.30E+06$\pm$1.14E+07 \\
\cline{2-6}
\multicolumn{1}{c|}{} & $cf_{30}$ & 2.06E+05$\pm$1.89E+05$-$ & 1.13E+04$\pm$2.43E+04 & 1.37E+05$\pm$1.12E+05$-$ & 3.59E+03$\pm$1.73E+03 \\
\hline
\multicolumn{2}{c|}{$+$}   &  \multicolumn{2}{c||}{1} & \multicolumn{ 2}{c}{3} \\  \hline
\multicolumn{2}{c|}{$-$}   & \multicolumn{2}{c||}{27} & \multicolumn{ 2}{c}{24} \\  \hline
\multicolumn{2}{c|}{$\approx$} & \multicolumn{2}{c||}{2} & \multicolumn{ 2}{c}{3} \\  \hline
\end{tabular}
\end{table*}

\section {Experimental Study}\label{sec:ES}
In this section, our experiments were conducted on 30 test functions with 30 dimensions (30D) and 50 dimensions (50D) at IEEE CEC2014. These 30 test functions are denoted as $cf_{1}$-$cf_{30}$, and their details can be available from ~\cite{liang2013problem}. In general, these 30 test functions can be grouped into four classes: 1) Unimodal functions $cf_{1}$-$cf_{3}$; 2) Simple multimodal functions $cf_{4}$-$cf_{16}$; 3) Hybrid functions $cf_{17}$-$cf_{22}$; and 4) Composition functions $cf_{23}$-$cf_{30}$.

In our experiments, $51$ independent runs were carried out for each test function. A run will terminate if the maximum number of fitness evaluations (FEs) is reached, which was recommended to be $10000*D$ ~\cite{liang2013problem}. At the end of a run, the function error value $(f(\vec{x}^{best})-f(\vec{x}^{*}))$  was recorded, where $\vec{x}^{*}$ is the optimal solution and  $\vec{x}^{best}$ denotes the searched best solution. If the function error value is less than $10^{-8}$, it was taken as zero. The average and standard deviation of the function error values in all runs (denoted as ``Mean Error'' and ``Std Dev'') were used to measure the performance of an algorithm. Besides, to test the statistical significance of the experimental results between two algorithms, the Wilcoxon's rank sum test at a 0.05 significance level was performed. There are two new parameters in ACoS: the archive size $AS$ and the punishment coefficient $\eta$, which were fixed to be $3 \times NP$ and $0.1$ in all simulations, respectively.

For the sake of convenience, if a specific EA is under the framework of ACoS, the name of this EA will be modified by adding four letters ``ACoS-''. For example, PSO-w under our framework is named as ACoS-PSO-w.

\begin{table*} [!t]
\scriptsize
\centering
\addtolength{\tabcolsep}{-4pt}
\renewcommand{\arraystretch}{1.25}
\caption{Experimental results of PSO-w, ACoS-PSO-w, PSO-cf, and ACoS-PSO-cf over 51 independent runs on 30 test functions with 50D from IEEE CEC2014 using 500,000 FEs.}\label{tb2:pso50D}
\begin{threeparttable}[b]
\newcommand{\minitab}[2][l]{\begin{tabular}{#1}#2\end{tabular}}
\newcommand{\tabincell}[2]{\begin{tabular}{@{}#1@{}}#2\end{tabular}}
\begin{tabular}[width=\linewidth]{r|c|c|c||c|c}
\hline
\hline
\multicolumn{2}{c|}{\multirow{2}{*}{\tabincell{c}{Test Functions with 50D\\ from IEEE CEC2014}}} & PSO-w & ACoS-PSO-w & PSO-cf & ACoS-PSO-cf \\
\multicolumn{2}{c|}{} & Mean Error$\pm$Std Dev & Mean Error$\pm$Std Dev & Mean Error ±~Std Dev & Mean Error$\pm$Std Dev \\
\hline
\multicolumn{1}{c|}{\multirow{3}{*}{\tabincell{c}{Unimodal\\Functions}}} & $cf_{1}$  & 5.23E+08$\pm$2.85E+08$-$ & 6.60E+06$\pm$7.39E+06 & 3.30E+08$\pm$2.76E+08$-$ & 1.23E+05$\pm$8.87E+04 \\
\cline{2-6}
\multicolumn{1}{c|}{} & $cf_{2}$  & 5.62E+10$\pm$1.57E+10$-$ & 1.48E+08$\pm$7.73E+08 & 3.06E+10$\pm$1.31E+10$-$ & 1.91E+03$\pm$4.55E+03 \\
\cline{2-6}
\multicolumn{1}{c|}{} & $cf_{3}$  & 7.77E+04$\pm$3.61E+04$-$ & 1.76E+02$\pm$5.88E+02 & \textbf{1.42E+04$\pm$1.81E+04$-$} & \textbf{0.00E+00$\pm$0.00E+00} \\
\cline{1-6}
\multicolumn{1}{c|}{\multirow{13}{*}{\tabincell{c}{Simple\\Multimodal\\ Functions}}} & $cf_{4}$ & 6.92E+03$\pm$3.34E+03$-$ & 1.63E+02$\pm$1.75E+02 & 2.24E+03$\pm$1.75E+03$-$ & 9.10E+01$\pm$3.14E+01 \\
\cline{2-6}
\multicolumn{1}{c|}{} & $cf_{5}$  & 2.10E+01$\pm$8.68E-02$\approx$ & 2.10E+01$\pm$8.98E-02 & 2.02E+01$\pm$2.65E-01$+$ & 2.10E+01$\pm$1.44E-01 \\
\cline{2-6}
\multicolumn{1}{c|}{} & $cf_{6}$ & 4.46E+01$\pm$5.13E+00$-$ & 4.11E+01$\pm$4.91E+00 & 4.10E+01$\pm$4.75E+00$-$ & 3.87E+01$\pm$5.60E+00 \\
\cline{2-6}
\multicolumn{1}{c|}{} & $cf_{7}$ & 4.89E+02$\pm$1.52E+02$-$ & 1.14E+01$\pm$1.91E+01 & 2.64E+02$\pm$1.27E+02$-$ & 4.92E-03$\pm$7.13E-03 \\
\cline{2-6}
\multicolumn{1}{c|}{} & $cf_{8}$ & 2.75E+02$\pm$3,86E+01$-$ & 1.79E+02$\pm$3.43E+01 & 2.23E+02$\pm$5.38E+01$-$ & 1.75E+02$\pm$3.69E+01 \\
\cline{2-6}
\multicolumn{1}{c|}{} & $cf_{9}$ & 3.60E+02$\pm$6.32E+01$-$ & 2.37E+02$\pm$5.10E+01 & 3.18E+02$\pm$7.13E+01$-$ & 2.03E+02$\pm$5.37E+01 \\
 \cline{2-6}
\multicolumn{1}{c|}{} & $cf_{10}$  & 6.95E+03$\pm$1.04E+03$-$ & 5.76E+03$\pm$1.02E+03 & 5.72E+03$\pm$1.11E+03$+$ & 6.34E+03$\pm$8.75E+02 \\
\cline{2-6}
\multicolumn{1}{c|}{} & $cf_{11}$ & 7.21E+03$\pm$9.63E+02$-$ & 7.02E+03$\pm$1.17E+03 & 6.91E+03$\pm$9.82E+02$-$ & 6.67E+03$\pm$9.22E+02 \\
\cline{2-6}
\multicolumn{1}{c|}{} & $cf_{12}$ & 8.91E-01$\pm$6.81E-01$+$ & 1.46E+00$\pm$8.14E-01 & 4.64E-01$\pm$1.47E-01$+$ & 9.89E-01$\pm$6.14E-01 \\
 \cline{2-6}
\multicolumn{1}{c|}{} & $cf_{13}$ & 4.85E+00$\pm$8.24E-01$-$ & 8.10E-01$\pm$1.06E-01 & 3.50E+00$\pm$1.20E+00$-$ & 5.19E-01$\pm$9.60E-02 \\
\cline{2-6}
\multicolumn{1}{c|}{} & $cf_{14}$ & 1.33E+02$\pm$3.27E+01$-$ & 1.49E+00$\pm$2.53E-01 & 7.60E+01$\pm$3.75E+01$-$ & 5.63E-01$\pm$2.74E-01 \\
\cline{2-6}
\multicolumn{1}{c|}{} & $cf_{15}$ & 4.54E+05$\pm$5.86E+05$-$ & 1.82E+03$\pm$6.01E+03 & 1.65E+05$\pm$3.04E+05$-$ & 1.02E+01$\pm$2.70E+00 \\
\cline{2-6}
\multicolumn{1}{c|}{} & $cf_{16}$ & 2.07E+01$\pm$7.28E-01$\approx$ & 2.08E+01$\pm$7.38E-01 & 2.07E+01$\pm$8.18E-01$\approx$ & 2.07E+01$\pm$5.94E-01 \\
\cline{1-6}
\multicolumn{1}{c|}{\multirow{6}{*}{\tabincell{c}{Hybrid\\Functions}} }& $cf_{17}$ & 2.89E+07$\pm$2.86E+07$-$ & 3.29E+05$\pm$4.22E+05 & 1.50E+07$\pm$2.04E+07$-$ & 3.51E+03$\pm$2.73E+03 \\
\cline{2-6}
\multicolumn{1}{c|}{} & $cf_{18}$ & 1.42E+09$\pm$1.12E+09$-$ & 1.95E+03$\pm$1.61E+03 & 1.02E+09$\pm$9.81E+08$-$ & 4.23E+03$\pm$1.89E+03 \\
\cline{2-6}
\multicolumn{1}{c|}{} & $cf_{19}$ & 3.53E+02$\pm$1.82E+02$-$ & 6.41E+01$\pm$3.97E+01 & 2.89E+02$\pm$2.52E+02$-$ & 2.52E+01$\pm$9.93E+00 \\
\cline{2-6}
\multicolumn{1}{c|}{} & $cf_{20}$ & 5.37E+04$\pm$3.90E+04$-$ & 9.46E+02$\pm$5.10E+02 & 7.94E+03$\pm$1.40E+04$-$ & 6.09E+02$\pm$1.62E+02 \\
\cline{2-6}
\multicolumn{1}{c|}{} & $cf_{21}$ & 9.21E+06$\pm$1.10E+07$-$ & 2.59E+05$\pm$4.70E+05 & 5.14E+06$\pm$8.56E+06$-$ & 2.11E+03$\pm$7.27E+02 \\
\cline{2-6}
\multicolumn{1}{c|}{} & $cf_{22}$ & 1.57E+03$\pm$4.43E+02$-$ & 1.10E+03$\pm$3.12E+02 & 1.54E+03$\pm$3.85E+02$-$ & 1.04E+03$\pm$2.97E+02 \\
\cline{1-6}
\multicolumn{1}{c|}{\multirow{8}{*}{\tabincell{c}{Composition\\Functions }}} & $cf_{23}$ & 6.62E+02$\pm$1.45E+02$-$ & 3.49E+02$\pm$1.87E+01 & 5.29E+02$\pm$1.16E+02$-$ & 3.44E+02$\pm$4.69E-13 \\
\cline{2-6}
\multicolumn{1}{c|}{} & $cf_{24}$ & 4.04E+02$\pm$4.57E+01$-$ & 2.99E+02$\pm$1.83E+01 & 3.49E+02$\pm$3.44E+01$-$ & 2.91E+02$\pm$6.15E+00 \\
\cline{2-6}
\multicolumn{1}{c|}{} & $cf_{25}$ & 2.52E+02$\pm$2.15E+01$-$ & 2.13E+02$\pm$5.08E+00 & 2.29E+02$\pm$1.33E+01$-$ & 2.17E+02$\pm$9.18E+00 \\
\cline{2-6}
\multicolumn{1}{c|}{} & $cf_{26}$ & 1.86E+02$\pm$1.19E+02$\approx$ & 1.88E+02$\pm$1.18E+02  & 1.87E+02$\pm$9.95E+01$-$ & 1.32E+02$\pm$9.67E+01 \\
\cline{2-6}
\multicolumn{1}{c|}{} & $cf_{27}$ & 1.75E+03$\pm$1.22E+02$-$ & 1.52E+03$\pm$1.42E+02 & 1.62E+03$\pm$1.80E+02$-$ & 1.51E+03$\pm$2.09E+02 \\
\cline{2-6}
\multicolumn{1}{c|}{} & $cf_{28}$ & 3.68E+03$\pm$8.30E+02$-$ & 2.42E+03$\pm$5.84E+02 & 3.73E+03$\pm$9.88E+02$-$ & 2.93E+03$\pm$8.20E+02 \\
\cline{2-6}
\multicolumn{1}{c|}{} & $cf_{29}$ & 1.23E+08$\pm$5.07E+07$-$ & 7.00E+07$\pm$4.64E+07 & 1.25E+08$\pm$7.59E+07$-$ & 3.88E+07$\pm$4.56E+07 \\
\cline{2-6}
\multicolumn{1}{c|}{} & $cf_{30}$ & 7.35E+05$\pm$6.25E+05$-$ & 1.96E+04$\pm$1.16E+03 & 5.77E+05$\pm$4.36E+05$-$ & 1.43E+04$\pm$2.44E+03 \\
\hline
\multicolumn{2}{c|}{$+$}   & \multicolumn{2}{c||}{1} & \multicolumn{ 2}{c}{3} \\  \hline
\multicolumn{2}{c|}{$-$}   & \multicolumn{2}{c||}{26} & \multicolumn{ 2}{c}{26} \\  \hline
\multicolumn{2}{c|}{ $\approx$} & \multicolumn{2}{c||}{3} & \multicolumn{ 2}{c}{1} \\  \hline
\end{tabular}
\end{threeparttable}
\end{table*}

\subsection {ACoS for two popular PSO variants}
Firstly, we applied ACoS to two of the most popular PSO variants: PSO-w and PSO-cf, which have been introduced in Section~\ref{sec:PSO}. The resultant methods are denoted as ACoS-PSO-w and ACoS-PSO-cf, respectively.

The population size of these two PSO variants and their augmented algorithms was set to be 40 and 60 when the dimension of the search space was equal to $30$ and $50$, respectively. The experimental results on $cf_{1}$-$cf_{30}$ with 30D and 50D are given in Tables I-II, where ``$+$'', ``$-$'', and ``$\approx$'' denote that PSO-w or PSO-cf performs better than, worse than, and similar to its augmented algorithm, respectively. The last three rows of Tables I-II summarize the experimental results.

Important observations can be obtained from Tables I-II:
  \begin{itemize}
    \item In the case of $D=30$, ACoS-PSO-w and ACoS-PSO-cf have an edge over their original algorithms on 27 and 24 test functions, respectively. With respect to $D=50$, both ACoS-PSO-w and ACoS-PSO-cf achieve better performance than their original algorithms on 26 test functions. However, PSO-w and PSO-cf cannot surpass their augmented algorithms on more than three test functions when $D=30$ and $D=50$.
    \item ACoS-PSO-w and ACoS-PSO-cf are never inferior to their original algorithms on any unimodal functions, hybrid functions, and composition functions, regardless of the number of the decision variables.
    \item ACoS is able to achieve great performance improvement toward PSO-w and PSO-cf on all the unimodal functions (i.e., $cf_{1}$-$cf_{3}$), five simple modal functions (i.e., $cf_{4}$, $cf_{7}$, and $cf_{13}$-$cf_{15}$), four hybrid functions (i.e., $cf_{17}$, $cf_{18}$, $cf_{20}$, and $cf_{21}$), and two composition functions (i.e., $cf_{29}$ and $cf_{30}$). Moreover, ACoS offers the optimal solutions for three cases in all runs, which have been highlighted in \textbf{boldface} in Tables 1-2.
    \item It seems that the increase of the dimension (i.e., from $D=30$ to $D=50$) does not have a remarkable influence on the performance improvement of our framework.
  \end{itemize}

From the above observations, our framework significantly improves the performance of these two popular PSO variants, which indicates that: 1) there is a  necessity to consider both the Eigen and original coordinate systems in the design of PSO variants, and 2) the adaptive scheme in ACoS is capable of effectively utilizing these two coordinate systems. The convergence graphs of the average function error values derived from these two PSO variants and their augmented algorithms are plotted in Fig. 2 for two test functions (i.e., $cf_{1}$ with 30D and $cf_{18}$ with 30D).


\begin{figure*} [!t]
    \begin{center}
      \subfigure[$cf_{1}$ with 30D]{\includegraphics[width=5.0cm]{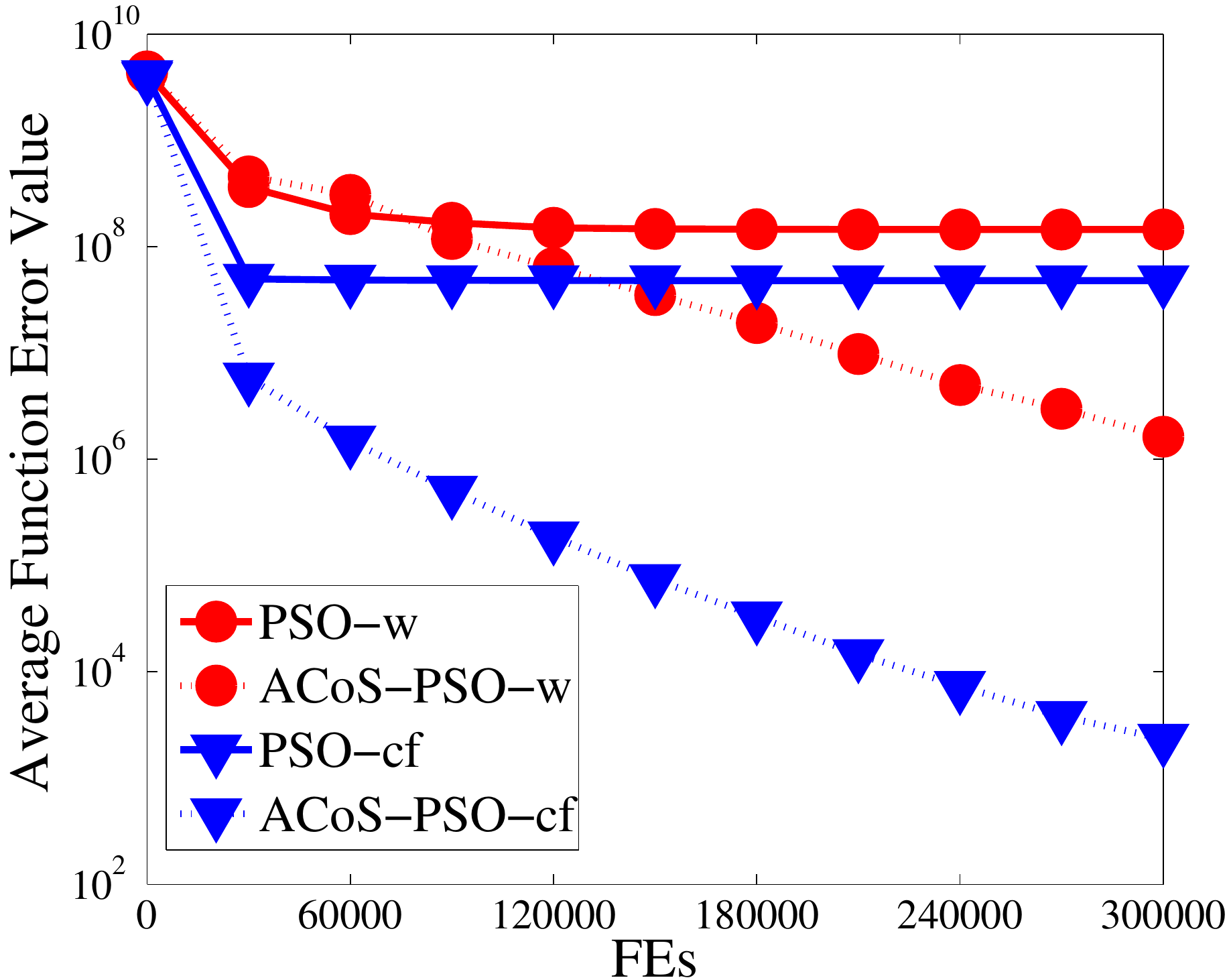}}
        \subfigure[$cf_{18}$ with 30D]{\includegraphics[width=5.0cm]{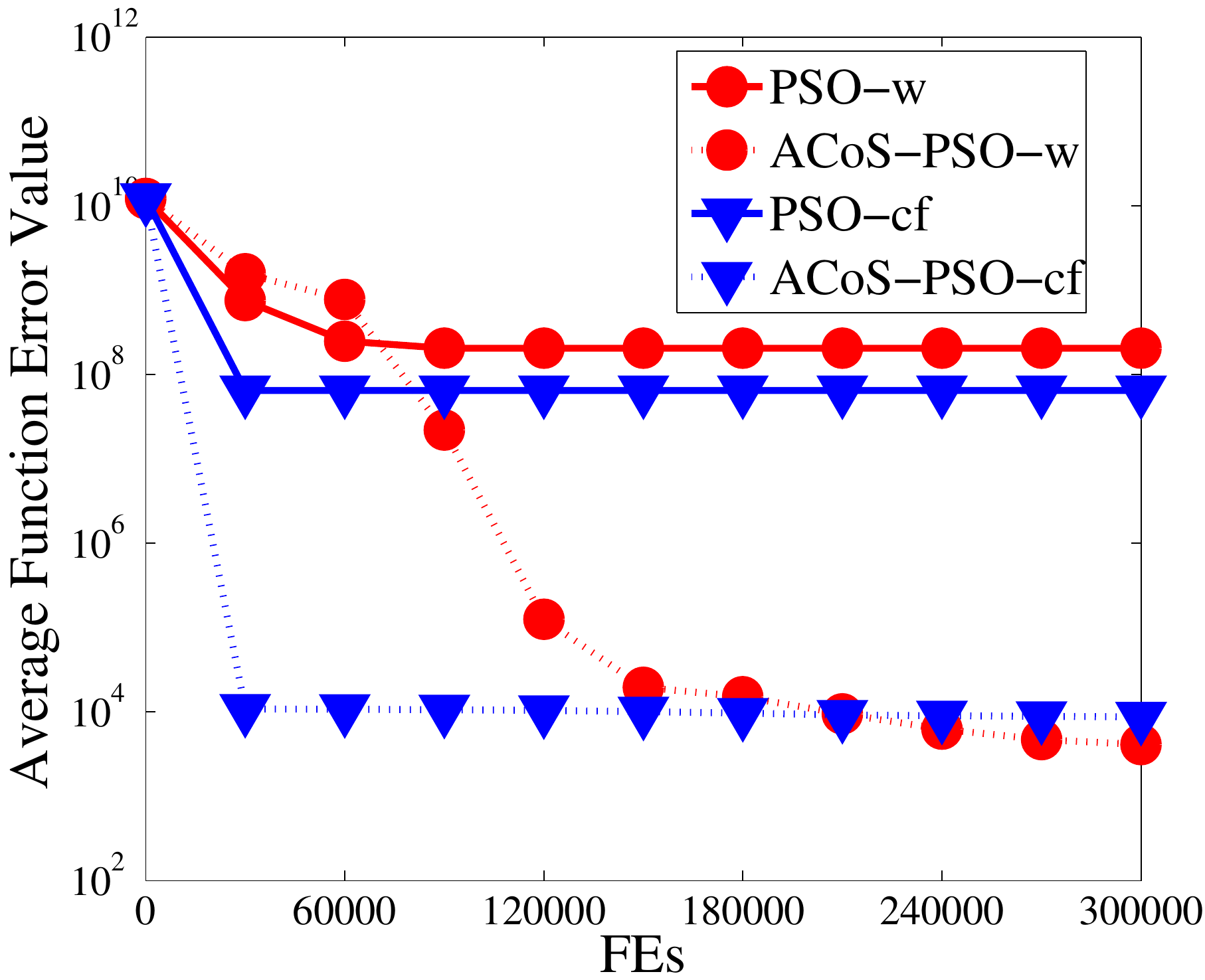}}
       \caption{Evolution of the average function error values derived from two popular PSO versions (PSO-w and PSO-cf ) and their augmented algorithms versus the number of FEs on $cf_{1}$ with 30D and $cf_{18}$ with 30D}
    \end{center} \label{fig:pso}
\end{figure*}

\subsection {ACoS for three state-of-the-art DE variants}
subsequently, we investigated the influence of ACoS on three famous DE variants: JADE, jDE, and SaDE. To ensure the comparison fair, the parameter settings of JADE, jDE, and SaDE were identical with their original papers, and remained unchanged when they were under the framework of ACoS. Tables III-IV show the comparison results on $cf_{1}$-$cf_{30}$ with 30D and 50D, where ``$+$'', ``$-$'', and``$\approx$'' denote that  a state-of-the-art DE variant performs better than, worse than, and similar to its augmented algorithm, respectively. The last three rows of Tables 5-6 summarize the results.

As can be seen from Tables III-IV, ACoS significantly improves JADE, jDE, and SaDE on many test functions. Specifically, compared with their original algorithms, when $D=30$, ACoS-JADE, ACoS-jDE and ACoS-SaDE obtain significance on 17, 13, and 21 test functions, respectively; meanwhile in the case of $D=50$, they outperform on 12, 14, and 23 test functions, respectively. In contrast, JADE, jDE, and SaDE cannot beat their augmented algorithms on more than four test functions. Besides, under our framework, these three state-of-the-art DE variants can consistently solve 14 cases, which have been highlighted in \textbf{boldface} in Tables 5-6. Moreover, the superiority of ACoS-jDE and ACoS-SaDE over their original algorithms increases as the dimension of the search space increases (i.e., from $D=30$ to $D=50$).

\begin{sidewaystable*}
\scriptsize
\centering
\addtolength{\tabcolsep}{-2pt}
\renewcommand{\arraystretch}{1.25}
\caption{Experimental results of JADE, ACoS-JADE, jDE, ACoS-jDE, SaDE and ACoS-SaDE over 51 independent runs on 30 test functions with 30D from IEEE CEC2014 using 300,000 FEs.}\label{tb2:famousDE30D}
\newcommand{\minitab}[2][l]{\begin{tabular}{#1}#2\end{tabular}}
\newcommand{\tabincell}[2]{\begin{tabular}{@{}#1@{}}#2\end{tabular}}
\begin{tabular}[width=\linewidth]{r|c|c|c||c|c||c|c}
\hline
\hline
\multicolumn{2}{c|}{\multirow{2}{*}{\tabincell{c}{Test Functions with 30D\\ from IEEE CEC2014}}} & JADE & ACoS-JADE & jDE & ACoS-jDE & SaDE & ACoS-SaDE \\

\multicolumn{2}{c|}{} & Mean Error$\pm$Std Dev & Mean Error$\pm$Std Dev & Mean Error$\pm$Std Dev & Mean Error$\pm$Std Dev& Mean Error$\pm$Std Dev & Mean Error$\pm$Std Dev \\
\hline
\multicolumn{1}{c|}{\multirow{3}{*}{\tabincell{c}{Unimodal\\Functions}}} & $cf_{1}$& \textbf{6.09E+02$\pm$1.18E+03}$-$ & \textbf{0.00E+00$\pm$0.00E+00} & \textbf{7.35E+04$\pm$6.12E+04}$-$ & \textbf{0.00E+00$\pm$0.00E+00} & \textbf{3.60E+05$\pm$2.74E+05}$-$ & \textbf{0.00E+00$\pm$0.00E+00} \\
\cline{2-8}
\multicolumn{1}{c|}{} & $cf_{2}$ & 0.00E+00$\pm$0.00E+00$\approx$ & 0.00E+00$\pm$0.00E+00 & 0.00E+00$\pm$0.00E+00$\approx$ & 0.00E+00$\pm$0.00E+00 &0.00E+00$\pm$0.00E+00$\approx$ & 0.00E+00$\pm$0.00E+00 \\
\cline{2-8}
\multicolumn{1}{c|}{} & $cf_{3}$ & \textbf{9.86E-04$\pm$5.95E-03}$-$ & \textbf{0.00E+00$\pm$0.00E+00} & 0.00E+00$\pm$0.00E+00$\approx$ & 0.00E+00$\pm$0.00E+00 & \textbf{1.92E+01$\pm$5.60E+01}$-$ & \textbf{0.00E+00$\pm$0.00E+00} \\
\cline{1-8}
\multicolumn{1}{c|}{\multirow{13}{*}{\tabincell{c}{Simple\\Multimodal \\Functions}}} & $cf_{4}$ & 0.00E+00$\pm$0.00E+00$\approx$ & 0.00E+00$\pm$0.00E+00 & 5.09E+00$\pm$1.48E+01$-$ & 1.24E+00$\pm$8.87E+00 & \textbf{4.13E+01$\pm$3.65E+01}$-$ & \textbf{0.00E+00$\pm$0.00E+00} \\
\cline{2-8}
\multicolumn{1}{c|}{} & $cf_{5}$ & 2.03E+01$\pm$3.23E-02$\approx$ & 2.03E+01$\pm$6.08E-02 & 2.03E+01$\pm$3.80E-02$\approx$ & 2.03E+01$\pm$4.44E-02 & 2.05E+01$\pm$4.94E-02$\approx$ & 2.03E+01$\pm$3.50E-02 \\
\cline{2-8}
\multicolumn{1}{c|}{} & $cf_{6}$ & 9.15E+00$\pm$2.21E+00$-$ & 6.11E+00$\pm$3.54E+00 & 3.39E+00$\pm$3.97E+00$-$ & 1.32E+00$\pm$2.31E+00 & 4.86E+00$\pm$2.15E+00$-$ & 2.93E+00$\pm$3.65E+00 \\
\cline{2-8}
\multicolumn{1}{c|}{} & $cf_{7}$ & 0.00E+00$\pm$0.00E+00$\approx$ & 0.00E+00$\pm$0.00E+00 & 0.00E+00$\pm$0.00E+00$\approx$ & 0.00E+00$\pm$0.00E+00 &\textbf{ 1.12E-02$\pm$1.50E-02}$-$ & \textbf{0.00E+00$\pm$0.00E+00 }\\
\cline{2-8}
\multicolumn{1}{c|}{} & $cf_{8}$ & 0.00E+00$\pm$0.00E+00$\approx$ & 0.00E+00$\pm$0.00E+00 & 0.00E+00$\pm$0.00E+00$\approx$ & 0.00E+00$\pm$0.00E+00 & \textbf{5.85E-02$\pm$2.36E-01}$-$ & \textbf{0.00E+00$\pm$0.00E+00} \\
\cline{2-8}
\multicolumn{1}{c|}{} & $cf_{9}$ & 2.62E+01$\pm$4.96E+00$-$ & 2.47E+01$\pm$3.89E+00 & 4.40E+01$\pm$5.33E+00$\approx$ & 4.30E+01$\pm$5.65E+00 & 3.72E+01$\pm$8.61E+00$+$ & 4.58E+01$\pm$7.67E+00 \\
\cline{2-8}
\multicolumn{1}{c|}{} & $cf_{10}$ & 8.16E-03$\pm$1.18E-02$-$ & 4.89E-03$\pm$9.84E-03 & 1.22E-03$\pm$4.94E-03$\approx$ & 1.22E-03$\pm$4.94E-03 & 2.82E-01$\pm$4.35E-01$-$ & 4.32E-02$\pm$3.08E-02 \\
\cline{2-8}
\multicolumn{1}{c|}{} & $cf_{11}$ & 1.67E+03$\pm$2.13E+02$\approx$ & 1.69E+03$\pm$1.88E+02 & 2.41E+03$\pm$3.11E+02$\approx$ & 2.55E+03$\pm$3.02E+02 & 3.25E+03$\pm$5.37E+02$-$ & 2.62E+03$\pm$2.87E+02 \\
\cline{2-8}
\multicolumn{1}{c|}{} & $cf_{12}$ & 2.67E-01$\pm$3.57E-02$\approx$ & 2.80E-01$\pm$3.76E-02 & 4.56E-01$\pm$6.46E-02$+$ & 4.92E-01$\pm$6.51E-02 & 7.95E-01$\pm$9.96E-02$-$ & 5.13E-01$\pm$6.75E-02 \\
\cline{2-8}
\multicolumn{1}{c|}{} & $cf_{13}$ & 2.20E-01$\pm$3.25E-02$\approx$ & 2.21E-01$\pm$3.64E-02 & 3.04E-01$\pm$3.54E-02$-$ & 2.82E-01$\pm$3.55E-02 & 2.66E-01$\pm$4.05E-02$+$ & 2.99E-01$\pm$4.94E-02 \\
\cline{2-8}
\multicolumn{1}{c|}{} & $cf_{14}$ & 2.41E-01$\pm$3.18E-02$-$ & 2.24E-01$\pm$3.04E-02 & 2.83E-01$\pm$2.95E-02$-$ & 2.63E-01$\pm$3.44E-02 & 2.35E-01$\pm$3.70E-02$\approx$ & 2.31E-01$\pm$2.73E-02 \\
\cline{2-8}
\multicolumn{1}{c|}{} & $cf_{15}$ & 3.20E+00$\pm$4.55E-01$\approx$ & 3.16E+00$\pm$3.76E-01 & 5.89E+00$\pm$7.23E-01$\approx$ & 5.88E+00$\pm$8.14E-01 & 4.10E+00$\pm$1.40E+00$+$ & 6.41E+00$\pm$8.18E-01 \\
\cline{2-8}
\multicolumn{1}{c|}{} & $cf_{16}$ & 9.30E+00$\pm$4.61E-01$\approx$ & 9.41E+00$\pm$4.33E-01 & 9.85E+00$\pm$3.81E-01$\approx$ & 9.99E+00$\pm$2.49E-01 & 1.10E+01$\pm$2.64E-01$-$ & 1.01E+01$\pm$3.12E-01 \\
\cline{1-8}
\multicolumn{1}{c|}{\multirow{6}{*}{\tabincell{c}{Hybrid\\Functions}} }& $cf_{17}$ & 1.91E+04$\pm$1.08E+05$-$ & 4.11E+02$\pm$1.56E+02 & 1.13E+03$\pm$9.03E+02$-$ & 2.86E+02$\pm$1.25E+02 & 1.40E+04$\pm$1.36E+04$-$ & 9.00E+02$\pm$4.25E+02 \\
\cline{2-8}
\multicolumn{1}{c|}{} & $cf_{18}$ & 1.14E+02$\pm$1.97E+02$-$ & 2.94E+01$\pm$1.95E+01 & 1.66E+01$\pm$6.53E+00$-$ & 1.39E+01$\pm$9.31E+00 & 3.52E+02$\pm$4.95E+02$-$ & 5.94E+01$\pm$3.10E+01 \\
\cline{2-8}
\multicolumn{1}{c|}{} & $cf_{19}$ & 4.48E+00$\pm$7.56E-01$\approx$ & 4.48E+00$\pm$7.83E-01 & 4.36E+00$\pm$5.94E-01$-$ & 4.16E+00$\pm$6.71E-01 & 6.31E+00$\pm$1.15E+01$-$ & 4.58E+00$\pm$7.50E-01 \\
\cline{2-8}
\multicolumn{1}{c|}{} & $cf_{20}$ & 3.11E+03$\pm$3.01E+03$-$ & 1.22E+01$\pm$4.65E+00 & 1.16E+01$\pm$3.51E+00$-$ & 8.51E+00$\pm$1.90E+00 & 1.39E+02$\pm$2.02E+02$-$ & 2.12E+01$\pm$1.08E+01 \\
\cline{2-8}
\multicolumn{1}{c|}{} & $cf_{21}$ & 1.33E+04$\pm$4.12E+04$-$ & 1.40E+02$\pm$1.00E+02 & 2.74E+02$\pm$1.71E+02$-$ & 1.01E+02$\pm$8.28E+01 & 4.46E+03$\pm$7.23E+03$-$ & 2.38E+02$\pm$1.24E+02 \\
\cline{2-8}
\multicolumn{1}{c|}{} & $cf_{22}$ & 1.44E+02$\pm$7.74E+01$-$ & 1.12E+02$\pm$7.46E+01 & 1.08E+02$\pm$7.15E+01$-$ & 7.21E+01$\pm$5.14E+01 & 1.54E+02$\pm$5.78E+01$-$ & 1.45E+02$\pm$8.03E+01 \\
\cline{1-8}
\multicolumn{1}{c|}{\multirow{8}{*}{\tabincell{c}{Composition\\Functions }}} & $cf_{23}$ & 3.15E+02$\pm$4.01E-13$\approx$ & 3.15E+02$\pm$3.59E-13 & 3.15E+02$\pm$4.01E-13$\approx$ & 3.15E+02$\pm$3.73E-13 & 3.15E+02$\pm$2.24E-13$\approx$ & 3.15E+02$\pm$2.45E-13 \\
\cline{2-8}
\multicolumn{1}{c|}{} & $cf_{24}$ & 2.25E+02$\pm$3.60E+00$\approx$ & 2.24E+02$\pm$1.85E+00 & 2.25E+02$\pm$2.56E+00$\approx$ & 2.23E+02$\pm$7.89E-01 & 2.26E+02$\pm$2.79E+00$\approx$ & 2.24E+02$\pm$9.64E-01 \\
\cline{2-8}
\multicolumn{1}{c|}{} & $cf_{25}$  & 2.03E+02$\pm$1.13E+00$\approx$ & 2.03E+02$\pm$4.16E-01 & 2.03E+02$\pm$5.31E-01$\approx$ & 2.02E+02$\pm$3.67E-01 & 2.08E+02$\pm$2.54E+00$\approx$ & 2.03E+02$\pm$1.62E+00 \\
\cline{2-8}
\multicolumn{1}{c|}{} & $cf_{26}$ & 1.02E+02$\pm$1.39E+01$\approx$ & 1.00E+02$\pm$4.27E-02 & 1.00E+02$\pm$4.02E-02$\approx$ & 1.00E+02$\pm$3.64E-02 & 1.11E+02$\pm$3.24E+01$-$ & 1.00E+02$\pm$4.15E-02 \\
\cline{2-8}
\multicolumn{1}{c|}{} & $cf_{27}$ & 3.35E+02$\pm$4.68E+01$\approx$ & 3.34E+02$\pm$4.61E+01 & 3.62E+02$\pm$4.69E+01$\approx$ & 3.70E+02$\pm$4.53E+01 & 4.20E+02$\pm$4.42E+01$-$ & 3.66E+02$\pm$4.28E+01 \\
\cline{2-8}
\multicolumn{1}{c|}{} & $cf_{28}$ & 7.96E+02$\pm$4.63E+01$\approx$ & 7.96E+02$\pm$4.57E+01 & 7.99E+02$\pm$2.68E+01$\approx$ & 8.01E+02$\pm$3.54E+01 & 8.93E+02$\pm$3.46E+01$-$ & 8.18E+02$\pm$4.69E+01 \\
\cline{2-8}
\multicolumn{1}{c|}{} & $cf_{29}$ & 8.28E+02$\pm$3.27E+02$-$ & 6.12E+02$\pm$2.00E+02 & 8.13E+02$\pm$7.12E+01$-$ & 5.43E+02$\pm$2.29E+02 & 1.10E+03$\pm$2.16E+02$-$ & 6.12E+02$\pm$1.67E+02 \\
\cline{2-8}
\multicolumn{1}{c|}{} & $cf_{30}$ & 1.66E+03$\pm$7.61E+02$-$ & 1.02E+03$\pm$4.21E+02 & 1.40E+03$\pm$5.06E+02$-$ & 7.83E+02$\pm$3.71E+02 & 1.48E+03$\pm$5.40E+02$-$ & 9.75E+02$\pm$4.91E+02 \\
\hline
\multicolumn{2}{c|}{$+$}   &  \multicolumn{2}{c||}{0} & \multicolumn{ 2}{c||}{1} & \multicolumn{ 2}{c}{3} \\  \hline
\multicolumn{2}{c|}{$-$}   & \multicolumn{2}{c||}{17} & \multicolumn{ 2}{c||}{13} & \multicolumn{ 2}{c}{21} \\  \hline
\multicolumn{2}{c|}{ $\approx$} & \multicolumn{2}{c||}{13} &  \multicolumn{2}{c||}{16} & \multicolumn{ 2}{c}{6} \\  \hline
\end{tabular}
\end{sidewaystable*}

\begin{sidewaystable*}
\scriptsize
\centering
\addtolength{\tabcolsep}{-3pt}
\renewcommand{\arraystretch}{1.25}
\caption{Experimental results of JADE, ACoS-JADE, jDE, ACoS-jDE, SaDE and ACoS-SaDE over 51 independent runs on 30 test functions with 50D from IEEE CEC2014 using 500,000 FEs.}\label{tb2:famousDE50D}
\newcommand{\minitab}[2][l]{\begin{tabular}{#1}#2\end{tabular}}
\newcommand{\tabincell}[2]{\begin{tabular}{@{}#1@{}}#2\end{tabular}}
\begin{tabular}[width=\linewidth]{r|c|c|c||c|c||c|c}
\hline
\hline
\multicolumn{2}{c|}{\multirow{2}{*}{\tabincell{c}{Test Functions with 50D\\ from IEEE CEC2014}}} & JADE & ACoS-JADE & jDE & ACoS-jDE & SaDE & ACoS-SaDE \\
\multicolumn{2}{c|}{} & Mean Error$\pm$Std Dev & Mean Error$\pm$Std Dev & Mean Error$\pm$Std Dev & Mean Error ±~Std Dev & Mean Error$\pm$Std Dev & Mean Error$\pm$Std Dev \\
\hline
\multicolumn{1}{c|}{\multirow{3}{*}{\tabincell{c}{Unimodal\\Functions}}} & $cf_{1}$ & \textbf{1.45E$\pm$04 ±~9.43E+03}$-$ & \textbf{0.00E+00$\pm$0.00E+00} & 4.58E+05$\pm$2.03E+05$-$ & 2.94E+02$\pm$1.30E+03 & 9.38E+05$\pm$2.96E+05$-$ & 4.63E+03 ±~6.03E+03 \\
\cline{2-8}
\multicolumn{1}{c|}{} & $cf_{2}$ & 0.00E+00$\pm$0.00E+00$\approx$ & 0.00E+00$\pm$0.00E+00 & \textbf{9.19E-09$\pm$1.83E-08}$-$ & \textbf{0.00E+00$\pm$0.00E+00} & 3.65E+03$\pm$4.02E+03$-$ & 2.90E+02 ±~6.17E+02 \\
\cline{2-8}
\multicolumn{1}{c|}{} & $cf_{3}$ & \textbf{3.96E+03$\pm$2.41E+03}$-$ & \textbf{0.00E+00$\pm$0.00E+00} & 0.00E+00$\pm$0.00E+00$\approx$ & 0.00E+00$\pm$0.00E+00 & \textbf{3.04E+03$\pm$1.64E+03}$-$ & \textbf{0.00E+00 ±~0.00E+00} \\
\cline{1-8}
\multicolumn{1}{c|}{\multirow{13}{*}{\tabincell{c}{Simple\\Multimodal \\Functions}}} & $cf_{4}$ & 2.35E+01$\pm$4.12E+01$-$ & 1.10E+01$\pm$3.07E+01 & 8.70E+01$\pm$1.92E+01$-$ & 4.45E+01$\pm$3.94E+01 & 9.34E+01$\pm$3.89E+01$-$ & 2.84E+01 ±~3.24E+01 \\
\cline{2-8}
\multicolumn{1}{c|}{} & $cf_{5}$ & 2.03E+01$\pm$2.81E-02$\approx$ & 2.03E+01$\pm$3.68E-02 & 2.04E+01$\pm$3.30E-02$\approx$ & 2.04E+01$\pm$3.39E-02 & 2.07E+01$\pm$4.62E-02$\approx$ & 2.05E+01 ±~3.40E-02 \\
\cline{2-8}
\multicolumn{1}{c|}{} & $cf_{6}$ & 1.56E+01$\pm$6.56E+00$-$ & 1.17E+01$\pm$7.19E+00 & 8.88E+00$\pm$7.14E+00$-$ & 6.34E+00$\pm$5.31E+00 & 1.77E+01$\pm$3.55E+00$-$ & 7.77E+00$\pm$2.56E+00 \\
\cline{2-8}
\multicolumn{1}{c|}{} & $cf_{7}$ & 2.84E-03$\pm$6.74E-03$-$ & 8.69E-04$\pm$2.73E-03 & 0.00E+00$\pm$0.00E+00$\approx$ & 0.00E+00$\pm$0.00E+00 & 1.43E-02$\pm$1.36E-02$-$ & 3.28E-03$\pm$4.56E-03 \\
\cline{2-8}
\multicolumn{1}{c|}{} & $cf_{8}$ & 0.00E+00$\pm$0.00E+00$\approx$ & 0.00E+00$\pm$0.00E+00 & 0.00E+00$\pm$0.00E+00$\approx$ & 0.00E+00$\pm$0.00E+00 & \textbf{1.46E+00$\pm$1.78E+00}$-$ & \textbf{0.00E+00$\pm$0.00E+00} \\
\cline{2-8}
\multicolumn{1}{c|}{} & $cf_{9}$ & 5.15E+01$\pm$7.85E+00$\approx$ & 5.28E+01$\pm$6.61E+00 & 9.15E+01$\pm$9.51E+00$-$ & 8.92E+01$\pm$1.12E+01 & 8.78E+01$\pm$1.45E+01$+$ & 1.01E+02$\pm$2.16E+01 \\
 \cline{2-8}
\multicolumn{1}{c|}{} & $cf_{10}$ & 1.17E-02$\pm$1.33E-02$-$ & 9.79E-03$\pm$1.09E-02 & 7.34E-04$\pm$2.96E-03$+$ & 2.69E-03$\pm$5.18E-03 & 1.57E+00$\pm$1.11E+00$-$ & 1.06E-01$\pm$1.35E-01 \\
\cline{2-8}
\multicolumn{1}{c|}{} & $cf_{11}$ & 3.84E+03$\pm$3.04E+02$\approx$ & 3.93E+03$\pm$3.76E+02 & 5.22E+03$\pm$3.62E+02$\approx$ & 5.26E+03$\pm$3.88E+02 & 6.49E+03$\pm$1.70E+03$-$ & 6.07E+03$\pm$4.09E+02 \\
\cline{2-8}
\multicolumn{1}{c|}{} & $cf_{12}$ & 2.50E-01$\pm$3.42E-02$\approx$ & 2.64E-01$\pm$3.71E-02 & 4.93E-01$\pm$5.40E-02$\approx$ & 5.02E-01$\pm$5.75E-02 & 1.10E+00$\pm$1.10E-01$-$ & 6.27E-01$\pm$8.33E-02 \\
 \cline{2-8}
\multicolumn{1}{c|}{} & $cf_{13}$ & 3.29E-01$\pm$4.25E-02$-$ & 3.16E-01$\pm$4.45E-02 & 3.84E-01$\pm$4.45E-02$-$ & 3.68E-01$\pm$3.80E-02 & 4.26E-01$\pm$5.82E-02$-$& 4.01E-01$\pm$6.71E-02 \\
\cline{2-8}
\multicolumn{1}{c|}{} & $cf_{14}$ & 3.04E-01$\pm$8.56E-02$\approx$ & 2.98E-01$\pm$9.03E-02 & 3.26E-01$\pm$5.50E-02$-$ & 2.95E-01$\pm$3.04E-02 & 3.09E-01$\pm$3.72E-02$-$ & 2.93E-01$\pm$2.95E-02 \\
\cline{2-8}
\multicolumn{1}{c|}{} & $cf_{15}$ & 7.17E+00$\pm$8.18E-01$+$ & 7.35E+00$\pm$8.60E-01 & 1.20E+01$\pm$1.28E+00$\approx$ & 1.19E+01$\pm$1.39E+00 & 1.46E+01$\pm$4.25E+00$+$ & 1.74E+01$\pm$2.32E+00 \\
\cline{2-8}
\multicolumn{1}{c|}{} & $cf_{16}$ & 1.77E+01$\pm$3.71E-01$\approx$ & 1.78E+01$\pm$5.19E-01 & 1.82E+01$\pm$3.72E-01$\approx$ & 1.84E+01$\pm$4.20E-01 & 2.01E+01$\pm$3.19E-01$-$ & 1.88E+01$\pm$3.95E-01 \\
\cline{1-8}
\multicolumn{1}{c|}{\multirow{6}{*}{\tabincell{c}{Hybrid\\Functions}} }& $cf_{17}$ & 2.40E+03$\pm$6.31E+02$\approx$ & 2.42E+03$\pm$5.03E+02 & 2.16E+04$\pm$1.32E+04$-$ & 2.09E+03$\pm$5.36E+02 & 5.72E+04$\pm$3.41E+04$-$ & 2.12E+03$\pm$5.72E+02 \\
\cline{2-8}
\multicolumn{1}{c|}{} & $cf_{18}$ & 1.74E+02$\pm$5.03E+01$\approx$ & 1.88E+02$\pm$4.92E+01 & 5.91E+02$\pm$7.38E+02$-$ & 1.22E+02$\pm$4.01E+01 & 5.82E+02$\pm$5.63E+02$-$ & 4.93E+02$\pm$4.00E+01 \\
\cline{2-8}
\multicolumn{1}{c|}{} & $cf_{19}$ & 1.30E+01$\pm$6.01E+00$-$ & 1.07E+01$\pm$3.80E+00 & 1.30E+01$\pm$4.48E+00$\approx$ & 1.20E+01$\pm$2.89E+00 & 1.39E+01$\pm$6.45E+00$+$ & 1.67E+01$\pm$1.04E+01 \\
\cline{2-8}
\multicolumn{1}{c|}{} & $cf_{20}$ & 8.27E+03$\pm$6.67E+03$-$ & 2.38E+02$\pm$7.09E+01 & 4.86E+01$\pm$1.64E+01$-$ & 4.43E+01$\pm$1.89E+01 & 8.79E+02$\pm$6.50E+02$-$ & 1.54E+02$\pm$7.20E+01 \\
\cline{2-8}
\multicolumn{1}{c|}{} & $cf_{21}$ & 1.25E+03$\pm$3.07E+02$+$ & 1.41E+03$\pm$4.21E+02 & 8.53E+03$\pm$7.94E+03$-$ & 9.80E+02$\pm$2.91E+02 & 6.25E+04$\pm$3.33E+04$-$ & 1.18E+03$\pm$5.07E+02 \\
\cline{2-8}
\multicolumn{1}{c|}{} & $cf_{22}$ & 4.81E+02$\pm$1.58E+02$-$ & 4.32E+02$\pm$1.54E+02 & 5.40E+02$\pm$1.12E+02$-$ & 4.30E+02$\pm$1.50E+02 & 4.80E+02$\pm$1.43E+02$+$ & 5.04E+02$\pm$1.08E+02 \\
\cline{1-8}
\multicolumn{1}{c|}{\multirow{8}{*}{\tabincell{c}{Composition\\Functions }}} & $cf_{23}$ & 3.44E+02$\pm$4.26E-13$\approx$ & 3.44E+02$\pm$4.98E-13 & 3.44E+02$\pm$4.17E-13$\approx$ & 3.44E+02$\pm$2.87E-13 & 3.44E+02$\pm$2.85E-13$\approx$ & 3.44E+02$\pm$2.87E-13 \\
\cline{2-8}
\multicolumn{1}{c|}{} & $cf_{24}$  & 2.74E+02$\pm$1.66E+00$\approx$ & 2.74E+02$\pm$2.20E+00 & 2.68E+02$\pm$2.17E+00$\approx$ & 2.67E+02$\pm$2.45E+00 & 2.75E+02$\pm$3.44E+00$\approx$ & 2.69E+02$\pm$4.68E+00 \\
\cline{2-8}
\multicolumn{1}{c|}{} & $cf_{25}$ & 2.16E+02$\pm$7.03E+00$-$ & 2.09E+02$\pm$2.64E+00 & 2.07E+02$\pm$1.47E+00$\approx$ & 2.07E+02$\pm$1.44E+00 & 2.17E+02$\pm$8.71E+00$-$ & 2.10E+02$\pm$9.31E+00 \\
\cline{2-8}
\multicolumn{1}{c|}{} & $cf_{26}$ & 1.00E+02$\pm$1.33E-01$\approx$ & 1.00E+02$\pm$6.88E-02 & 1.00E+02$\pm$3.69E-02$\approx$ & 1.00E+02$\pm$5.47E-02 & 1.94E+02$\pm$2.36E+01$-$ & 1.23E+02$\pm$4.26E+01 \\
\cline{2-8}
\multicolumn{1}{c|}{} & $cf_{27}$ & 4.48E+02$\pm$5.03E+01$+$ & 4.77E+02$\pm$6.60E+01 & 4.48E+02$\pm$7.13E+01$-$ & 4.04E+02$\pm$6.10E+01 & 7.67E+02$\pm$6.81E+01$-$ & 5.69E+02$\pm$6.85E+01 \\
\cline{2-8}
\multicolumn{1}{c|}{} & $cf_{28}$ & 1.18E+03$\pm$5.71E+01$\approx$ & 1.17E+03$\pm$9.34E+01 & 1.09E+03$\pm$3.35E+01$\approx$ & 1.13E+03$\pm$5.82E+01 & 1.41E+03$\pm$1.25E+02$-$ & 1.20E+03$\pm$6.72E+01 \\
\cline{2-8}
\multicolumn{1}{c|}{} & $cf_{29}$ & 9.00E+02$\pm$6.73E+01$-$ & 8.72E+02$\pm$8.50E+01 & 1.04E+03$\pm$1.95E+02$-$ & 8.22E+02$\pm$5.41E+01 & 1.43E+03$\pm$3.82E+02$-$ & 9.32E+02$\pm$1.06E+02 \\
\cline{2-8}
\multicolumn{1}{c|}{} & $cf_{30}$ & 9.71E+03$\pm$7.21E+02$\approx$ & 9.89E+03$\pm$1.08E+03 & 8.70E+03$\pm$4.77E+02$\approx$ & 8.88E+03$\pm$5.72E+02 & 1.19E+04$\pm$1.78E+03$-$ & 9.99E+03$\pm$9.46E+02 \\
\hline
\multicolumn{2}{c|}{$+$}   &  \multicolumn{2}{c||}{3} & \multicolumn{ 2}{c||}{1} & \multicolumn{ 2}{c}{4} \\  \hline
\multicolumn{2}{c|}{$-$}   & \multicolumn{2}{c||}{12} & \multicolumn{ 2}{c||}{14} & \multicolumn{ 2}{c}{23} \\  \hline
\multicolumn{2}{c|}{ $\approx$} & \multicolumn{2}{c||}{15} &  \multicolumn{2}{c||}{15} & \multicolumn{ 2}{c}{3} \\  \hline
\end{tabular}
\end{sidewaystable*}

The above comparison confirms that ACoS is an effective framework to improve the performance of these three state-of-the-art DE variants, which verifies the necessity to consider both the Eigen and original coordinate systems in an adaptive fashion when designing DE. Two convergence graphs are given in Fig. 3 for the performance comparison between these three state-of-the-art DE variants and their augmented algorithms.

\begin{figure*} [!t]
    \begin{center}
      \subfigure[$cf_{1}$ with 30D]{\includegraphics[width=5.0cm]{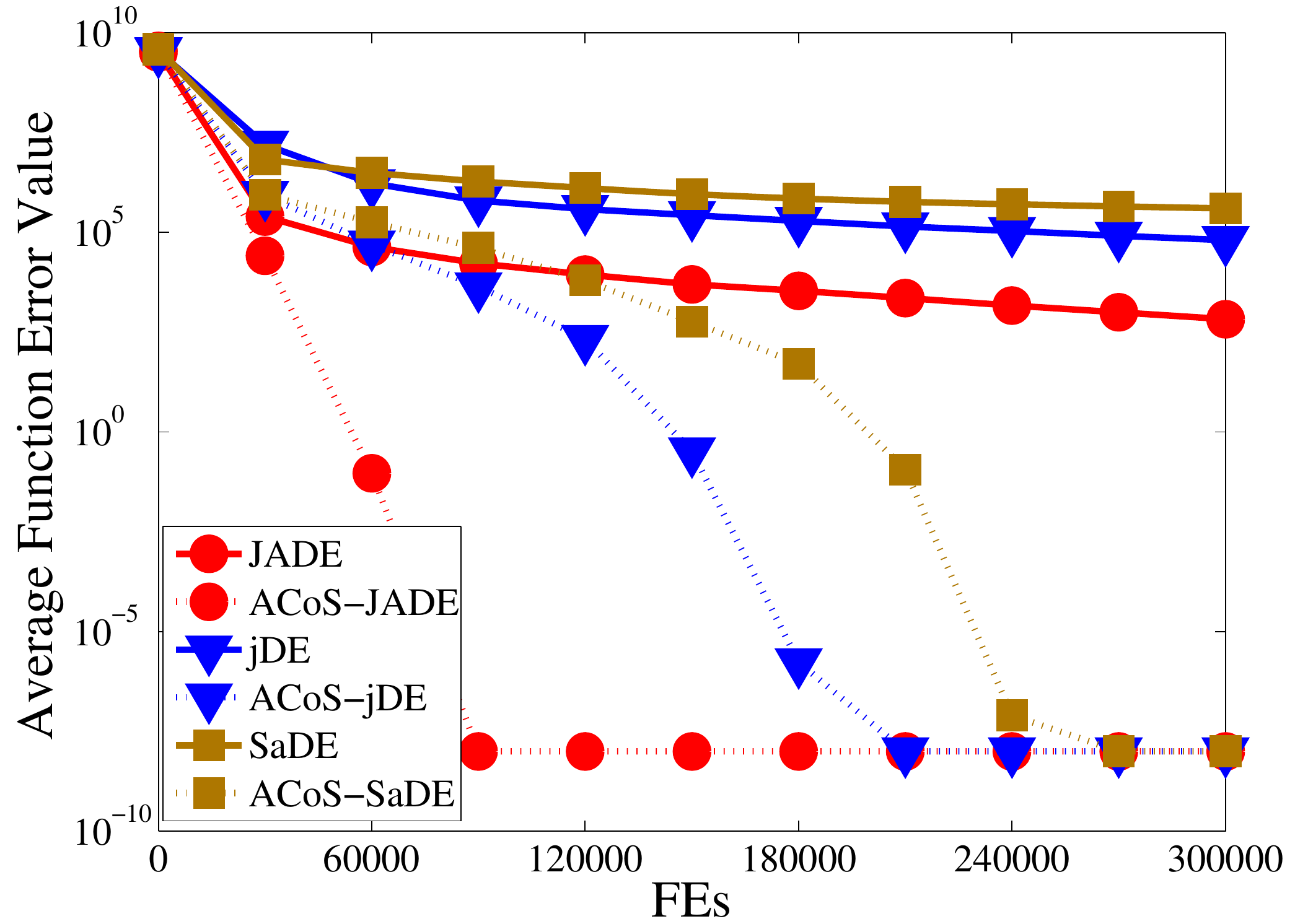}}
        \subfigure[$cf_{20}$ with 30D]{\includegraphics[width=5.0cm]{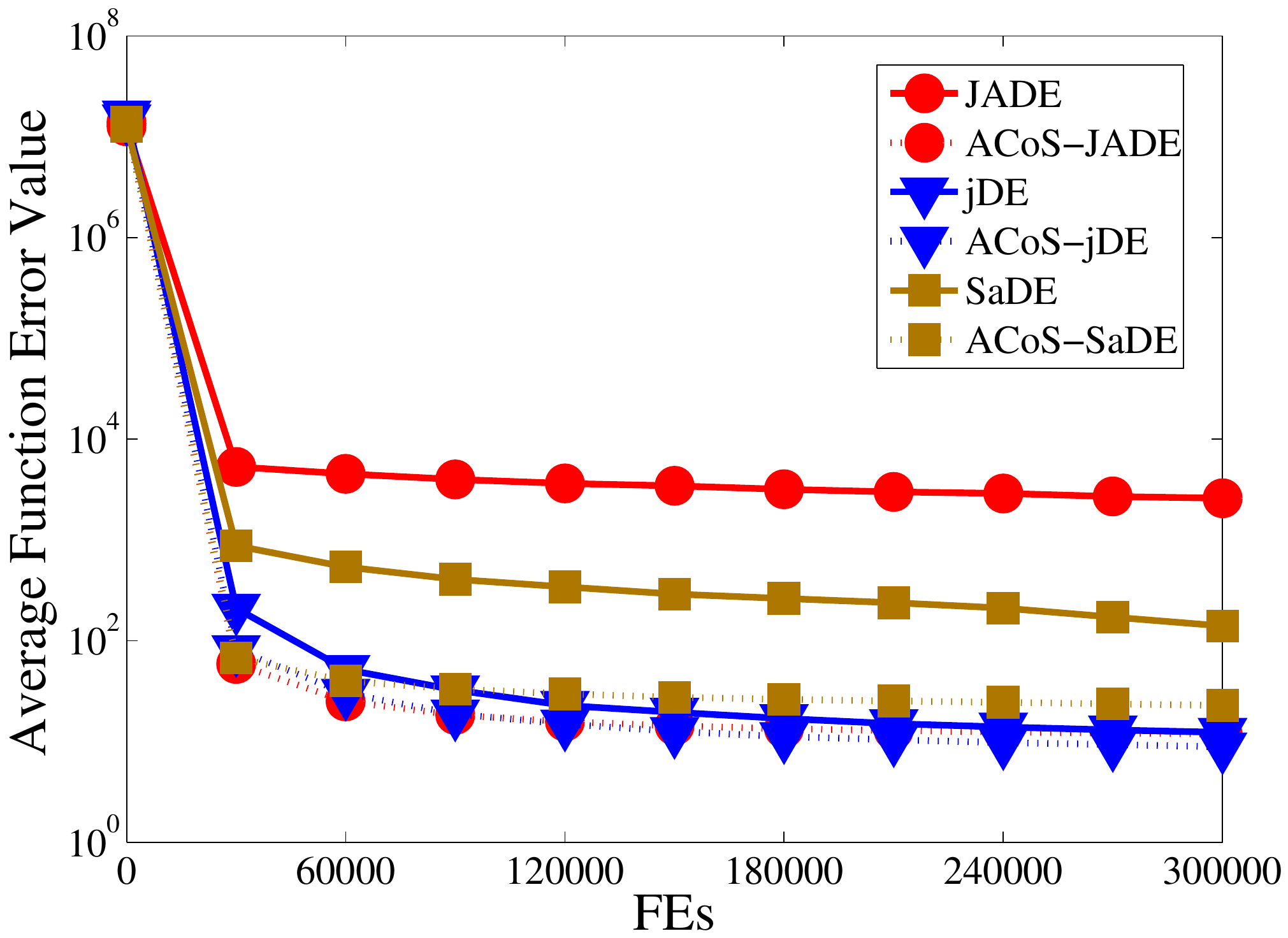}}
       \caption{Evolution of the average function error values derived from three state-of-the-art DE variants (JADE, jDE and SaDE ) and their augmented algorithms versus the number of FEs on $cf_{1}$ with 30D and $cf_{20}$ with 30D}
    \end{center} \label{fig:de}
\end{figure*}

\begin{table*}[!t]
\renewcommand{\arraystretch}{1.25}
\centering
\addtolength{\tabcolsep}{-4pt}
\scriptsize
\caption{Experimental results of CoJADE, JADE/eig, CPI-JADE, and ACoS-JADE over 51 independent runs on 30 test functions with 30D from IEEE CEC2014 using 300,000 FEs.}\label{tb2:withothermethods30D}
\newcommand{\minitab}[2][l]{\begin{tabular}{#1}#2\end{tabular}}
\newcommand{\tabincell}[2]{\begin{tabular}{@{}#1@{}}#2\end{tabular}}
\begin{tabular}[width=\linewidth]{r|c|c|c|c|c}
\hline
\hline
\multicolumn{2}{c|}{\multirow{2}{*}{\tabincell{c}{Test Functions with\\30D from IEEE CEC2014}}} & CoJADE &JADE/eig  &  CPI-JADE & ACoS-JADE \\
\multicolumn{2}{c|}{} & Mean Error$\pm$Std Dev & Mean Error$\pm$Std Dev & Mean Error$\pm$Std Dev & Mean Error$\pm$Std Dev \\
\hline
\multicolumn{1}{c|}{\multirow{3}{*}{\tabincell{c}{Unimodal\\Functions}}} & $cf_{1}$ & 3.96E+01$\pm$1.27E+02$-$ & 1.00E+02$\pm$3.12E+02$-$ & 0.00E+00$\pm$0.00E+00$\approx$ & 0.00E+00$\pm$0.00E+00 \\
\cline{2-6}
\multicolumn{1}{c|}{} & $cf_{2}$ & 0.00E+00$\pm$0.00E+00$\approx$ & 0.00E+00$\pm$0.00E+00$\approx$ & 0.00E+00$\pm$0.00E+00$\approx$ & 0.00E+00$\pm$0.00E+00 \\
\cline{2-6}
\multicolumn{1}{c|}{} & $cf_{3}$ & 5.90E-01$\pm$6.44E-01$-$ & 2.30E-02$\pm$3.83E-02$-$ & 0.00E+00$\pm$0.00E+00$\approx$ & 0.00E+00$\pm$0.00E+00 \\
\cline{1-6}
\multicolumn{1}{c|}{\multirow{13}{*}{\tabincell{c}{Simple\\Multimodal\\Functions}}} & $cf_{4}$ & 0.00E+00$\pm$0.00E+00$\approx$ & 0.00E+00$\pm$0.00E+00$\approx$ & 0.00E+00$\pm$0.00E+00$\approx$ & 0.00E+00$\pm$0.00E+00 \\
\cline{2-6}
\multicolumn{1}{c|}{} & $cf_{5}$ & 2.03E+01$\pm$1.09E-01$\approx$ & 2.03E+01$\pm$4.46E-02$\approx$ & 2.03E+01$\pm$3.68E-02$\approx$ & 2.03E+01$\pm$6.08E-02 \\
\cline{2-6}
\multicolumn{1}{c|}{} & $cf_{6}$ & 7.67E+00$\pm$3.51E+00$-$ & 6.98E+00$\pm$4.06E+00$-$ & 3.44E+00$\pm$3.57E+00$+$ & 6.11E+00$\pm$3.54E+00 \\
\cline{2-6}
\multicolumn{1}{c|}{} & $cf_{7}$ & 0.00E+00$\pm$0.00E+00$\approx$ & 0.00E+00$\pm$0.00E+00$\approx$ & 0.00E+00$\pm$0.00E+00$\approx$ & 0.00E+00$\pm$0.00E+00 \\
\cline{2-6}
\multicolumn{1}{c|}{} & $cf_{8}$ & 0.00E+00$\pm$0.00E+00$\approx$ & 0.00E+00$\pm$0.00E+00$\approx$ & 0.00E+00$\pm$0.00E+00$\approx$ & 0.00E+00$\pm$0.00E+00 \\
\cline{2-6}
\multicolumn{1}{c|}{} & $cf_{9}$ & 2.66E+01$\pm$4.25E+00$-$ & 2.46E+01$\pm$4.80E+00$\approx$ & 2.24E+01$\pm$5.33E+00$\approx$ & 2.47E+01$\pm$3.89E+00 \\
 \cline{2-6}
\multicolumn{1}{c|}{} & $cf_{10}$ & 7.57E-02$\pm$3.08E-02$-$ & 5.47E-01$\pm$1.55E-01$-$ & 3.83E-01$\pm$7.47E-02$-$ & 4.89E-03$\pm$9.84E-03 \\
\cline{2-6}
\multicolumn{1}{c|}{} & $cf_{11}$ & 1.76E+03$\pm$2.41E+02$\approx$ & 1.83E+03$\pm$2.25E+02$-$ & 1.77E+03$\pm$2.55E+02$\approx$ & 1.69E+03$\pm$1.88E+02 \\
\cline{2-6}
\multicolumn{1}{c|}{} & $cf_{12}$ & 2.93E-01$\pm$4.28E-02$-$ & 3.14E-01$\pm$5.91E-02$-$ & 3.95E-01$\pm$8.64E-02$-$ & 2.80E-01$\pm$3.76E-02 \\
 \cline{2-6}
\multicolumn{1}{c|}{} & $cf_{13}$ & 2.19E-01$\pm$3.64E-02$\approx$ & 2.23E-01$\pm$4.05E-02$\approx$ & 2.04E-01$\pm$3.38E-02$+$ & 2.21E-01$\pm$3.64E-02 \\
\cline{2-6}
\multicolumn{1}{c|}{} & $cf_{14}$ & 2.31E-01$\pm$3.02E-02$\approx$ & 2.31E-01$\pm$3.00E-02$-$ & 2.32E-01$\pm$3.35E-02$\approx$ & 2.24E-01$\pm$3.04E-02 \\
\cline{2-6}
\multicolumn{1}{c|}{} & $cf_{15}$ & 3.23E+00$\pm$3.59E-01$\approx$ & 3.27E+00$\pm$4.45E-01$\approx$ & 3.26E+00$\pm$3.78E-01$\approx$ & 3.16E+00$\pm$3.76E-01 \\
\cline{2-6}
\multicolumn{1}{c|}{} & $cf_{16}$ &9.62E+00$\pm$2.93E-01$-$ & 9.76E+00$\pm$3.48E-01$-$ & 9.70E+00$\pm$2.79E-01$-$ & 9.41E+00$\pm$4.33E-01 \\
\cline{1-6}
\multicolumn{1}{c|}{\multirow{6}{*}{\tabincell{c}{Hybrid\\Functions}} }& $cf_{17}$ & 1.26E+03$\pm$3.76E+02$-$ & 1.42E+03$\pm$4.38E+02$-$ & 1.16E+03$\pm$3.81E+02$-$ & 4.11E+02$\pm$1.56E+02 \\
\cline{2-6}
\multicolumn{1}{c|}{} & $cf_{18}$ & 1.01E+02$\pm$3.33E+01$-$ & 9.73E+01$\pm$3.70E+01$-$ & 9.47E+01$\pm$3.42E+01$-$ & 2.94E+01$\pm$1.95E+01 \\
\cline{2-6}
\multicolumn{1}{c|}{} & $cf_{19}$ & 4.66E+00$\pm$8.20E-01$-$ & 4.59E+00$\pm$6.63E-01$\approx$ & 4.89E+00$\pm$7.64E-01$-$ & 4.48E+00$\pm$7.83E-01 \\
\cline{2-6}
\multicolumn{1}{c|}{} & $cf_{20}$ & 5.55E+02$\pm$7.08E+02$-$ & 2.52E+02$\pm$2.57E+02$-$ & 1.12E+01$\pm$5.24E+00$\approx$ & 1.22E+01$\pm$4.65E+00 \\
\cline{2-6}
\multicolumn{1}{c|}{} & $cf_{21}$ & 1.45E+03$\pm$5.33E+03$-$ & 6.95E+02$\pm$1.11E+03$-$ & 3.33E+02$\pm$1.54E+02$-$ & 1.40E+02$\pm$1.00E+02 \\
\cline{2-6}
\multicolumn{1}{c|}{} & $cf_{22}$  & 1.07E+02$\pm$6.93E+01$\approx$ & 1.03E+02$\pm$6.94E+02$\approx$ & 9.99E+01$\pm$6.09E+01$\approx$ & 1.12E+02$\pm$7.46E+01 \\
\cline{1-6}
\multicolumn{1}{c|}{\multirow{8}{*}{\tabincell{c}{Composition\\Functions }}} & $cf_{23}$ & 3.15E+02$\pm$4.01E-13$\approx$ & 3.15E+02$\pm$4.01E-13$\approx$ & 3.15E+02$\pm$4.01E-13$\approx$ & 3.15E+02$\pm$3.59E-13 \\
\cline{2-6}
\multicolumn{1}{c|}{} & $cf_{24}$ & 2.24E+02$\pm$1.72E+00$\approx$ & 2.25E+02$\pm$3.53E+00$\approx$ & 2.24E+02$\pm$2.93E+00$\approx$ & 2.24E+02$\pm$1.85E+00 \\
\cline{2-6}
\multicolumn{1}{c|}{} & $cf_{25}$ & 2.03E+02$\pm$9.00E-01$\approx$ & 2.03E+02$\pm$6.41E-01$\approx$ & 2.03E+02$\pm$5.77E-01$\approx$ & 2.03E+02$\pm$4.16E-01 \\
\cline{2-6}
\multicolumn{1}{c|}{} & $cf_{26}$ & 1.00E+02$\pm$4.71E-02$\approx$ & 1.00E+02$\pm$4.53E-02$\approx$ & 1.00E+02$\pm$2.92E-02$\approx$ & 1.00E+02$\pm$4.27E-02 \\
\cline{2-6}
\multicolumn{1}{c|}{} & $cf_{27}$ & 3.41E+02$\pm$4.96E+01$\approx$ & 3.45E+02$\pm$4.85E+01$-$ & 3.53E+02$\pm$5.03E+01$-$ & 3.34E+02$\pm$4.61E+01 \\
\cline{2-6}
\multicolumn{1}{c|}{} & $cf_{28}$ & 8.02E+02$\pm$3.81E+01$\approx$ & 8.06E+02$\pm$3.69E+01$-$ & 8.02E+02$\pm$4.34E+01$\approx$ & 7.96E+02$\pm$4.57E+01 \\
\cline{2-6}
\multicolumn{1}{c|}{} & $cf_{29}$ & 7.31E+02$\pm$1.26E+01$-$ & 7.30E+02$\pm$3.79E+01$-$ & 8.13E+02$\pm$7.12E+01$-$ & 6.12E+02$\pm$2.00E+02 \\
\cline{2-6}
\multicolumn{1}{c|}{} & $cf_{30}$ & 1.77E+03$\pm$8.13E+02$-$ & 1.68E+03$\pm$7.22E+02$-$ & 1.40E+03$\pm$7.24E+02$-$ & 1.02E+03$\pm$4.21E+02 \\
\hline
\multicolumn{2}{c|}{$+$}        &  0  & 0 & 2 \\  \cline{1-5}
\multicolumn{2}{c|}{$-$}        &  14 & 16 & 10 \\  \cline{1-5}
\multicolumn{2}{c|}{ $\approx$} &  16 & 14 & 18  \\  \cline{1-5}
\end{tabular}
\end{table*}

\subsection {Comparison between ACoS with other Eigen coordinate system based methods}
The aim of this subsection is to compare ACoS with other Eigen coordinate system based methods: CoBiDE, DE/eig, and CPI-DE. Due to its outstanding performance, JADE was selected as the instance algorithm. Afterward, we applied ACoS, CoBiDE, DE/eig, and CPI-DE to JADE and obtained ACoS-JADE, CoJADE, JADE/eig, and CPI-JADE, respectively. For fair comparison, CoJADE, JADE/eig, and CPI-JADE adopted the same parameter settings of $F$, $CR$, and $NP$ with the original JADE, while the other parameter settings were identical with their own original papers. $cf_{1}$-$cf_{30}$ with 30D were employed in the comparative study, and Table V summarizes the experimental results, where``$+$'', ``$-$'', and ``$\approx$'' denote that the performance of the corresponding algorithm is better than, worse than, and similar to that of ACoS-JADE, respectively.

As shown in Table V, ACoS-JADE exhibits the best performance among the four compared methods. It outperforms CoJADE, JADE/eig, and CPI-JADE on 14, 16 and 10 test functions, respectively; while only loses on no more than two test functions. It is worth noting that ACoS-JADE is never inferior to the three competitors on any unimodal functions, hybrid functions, and composition functions. Compared with CoJADE and JADE/eig, CPI-JADE and ACoS-JADE reach significantly better performance, which demonstrates the potential of utilizing the cumulative population distribution information rather than the single population distribution information to estimate the Eigen coordinate system. Compared with CPI-JADE, ACoS-JADE's superior performance is largely attributed to the usage of the additional archiving mechanism and the probability vector $\vec{p}$.

\subsection {The Benefit of ACoS's Components}

\begin{sidewaystable*}
\scriptsize
\centering
\addtolength{\tabcolsep}{-0pt}
\renewcommand{\arraystretch}{1.25}

\caption{Experimental results of nonAr-JADE, JADE, half-ACoS-JADE, Eig-ACoS-JADE, and ACoS-JADE over 51 independent runs on 30 test functions with 30D from IEEE CEC2014 using 300,000 FEs.
}\label{tb2:twocomponents30D}
\newcommand{\minitab}[2][l]{\begin{tabular}{#1}#2\end{tabular}}
\newcommand{\tabincell}[2]{\begin{tabular}{@{}#1@{}}#2\end{tabular}}
\begin{tabular}[width=\linewidth]{r|c|c|c|c|c|c}
\hline
\hline
\multicolumn{2}{c|}{\multirow{2}{*}{\tabincell{c}{Test Functions with\\30D from IEEE CEC2014}}} & nonAr-JADE, &JADE  & half-ACoS-JADE & Eig-ACoS-JADE & ACoS-JADE \\
\multicolumn{2}{c|}{} & Mean Error$\pm$Std Dev & Mean Error$\pm$Std Dev & Mean Error$\pm$Std Dev & Mean Error$\pm$Std Dev & Mean Error$\pm$Std Dev \\
\hline
\multicolumn{1}{c|}{\multirow{3}{*}{\tabincell{c}{Unimodal\\Functions}}} & $cf_{1}$ & 0.00E+00$\pm$0.00E+00$\approx$ & 6.09E+02$\pm$1.18E+03$-$ & 0.00E+00$\pm$0.00E+00$\approx$ & 0.00E+00$\pm$0.00E+00$\approx$   & 0.00E+00$\pm$0.00E+00 \\
\cline{2-7}
\multicolumn{1}{c|}{} & $cf_{2}$& 0.00E+00$\pm$0.00E+00$\approx$ & 0.00E+00$\pm$0.00E+00$\approx$  & 0.00E+00$\pm$0.00E+00$\approx$ & 0.00E+00$\pm$0.00E+00$\approx$  & 0.00E+00$\pm$0.00E+00 \\
\cline{2-7}
\multicolumn{1}{c|}{} & $cf_{3}$& 0.00E+00$\pm$0.00E+00$\approx$ & 9.86E-04$\pm$5.95E-03$-$ & 0.00E+00$\pm$0.00E+00$\approx$ & 0.00E+00$\pm$0.00E+00$\approx$   & 0.00E+00$\pm$0.00E+00 \\
\cline{1-7}
\multicolumn{1}{c|}{\multirow{13}{*}{\tabincell{c}{Simple\\Multimodal\\Functions}}} & $cf_{4}$& 0.00E+00$\pm$0.00E+00$\approx$ & 0.00E+00$\pm$0.00E+00$\approx$ & 0.00E+00$\pm$0.00E+00$\approx$ & 0.00E+00$\pm$0.00E+00$\approx$   & 0.00E+00$\pm$0.00E+00 \\
\cline{2-7}
\multicolumn{1}{c|}{} & $cf_{5}$& 2.03E+01$\pm$4.03E-02$\approx$ & 2.03E+01$\pm$3.23E-02$\approx$ & 2.03E+01$\pm$6.08E-02$\approx$ & 2.07E+01$\pm$3.02E-01$-$   & 2.03E+01$\pm$6.08E-02 \\
\cline{2-7}
\multicolumn{1}{c|}{} & $cf_{6}$& 6.60E+00$\pm$3.66E+00$-$ & 9.15E+00$\pm$2.21E+00$-$& 6.70E+00$\pm$3.44E+00$-$  & 8.19E+00$\pm$4.90E+00$-$   & 6.11E+00$\pm$3.54E+00 \\
\cline{2-7}
\multicolumn{1}{c|}{} & $cf_{7}$ & 0.00E+00$\pm$0.00E+00$\approx$ & 0.00E+00$\pm$0.00E+00$\approx$ & 0.00E+00$\pm$0.00E+00$\approx$ & 9.17E-04$\pm$2.95E-03$-$   & 0.00E+00$\pm$0.00E+00 \\
\cline{2-7}
\multicolumn{1}{c|}{} & $cf_{8}$ & 0.00E+00$\pm$0.00E+00$\approx$ & 0.00E+00$\pm$0.00E+00$\approx$ & 0.00E+00$\pm$0.00E+00$\approx$ & 2.69E+01$\pm$5.05E+00$-$   & 0.00E+00$\pm$0.00E+00 \\
\cline{2-7}
\multicolumn{1}{c|}{} & $cf_{9}$ & 2.41E+01$\pm$3.40E+00$\approx$ & 2.62E+01$\pm$4.96E+00$-$ & 2.46E+01$\pm$4.42E+00$\approx$ & 2.79E+01$\pm$5.73E+00$-$   & 2.47E+01$\pm$3.89E+00 \\
 \cline{2-7}
\multicolumn{1}{c|}{} & $cf_{10}$ & 6.93E-03$\pm$9.91E-03$-$ & 8.16E-03$\pm$1.18E-02$-$ & 5.64E-01$\pm$1.46E-01$-$  & 2.35E+03$\pm$2.35E+02$-$  & 4.89E-03$\pm$9.84E-03 \\
\cline{2-7}
\multicolumn{1}{c|}{} & $cf_{11}$& 1.72E+03$\pm$2.41E+02$\approx$ & 1.67E+03$\pm$2.13E+02$\approx$ & 1.82E+03$\pm$2.19E+02$-$ & 2.75E+03$\pm$2.75E+02$-$   & 1.69E+03$\pm$1.88E+02 \\
\cline{2-7}
\multicolumn{1}{c|}{} & $cf_{12}$ & 2.85E-01$\pm$3.95E-02$\approx$ & 2.67E-01$\pm$3.57E-02$\approx$ & 2.95E-01$\pm$4.39E-02$-$ & 6.32E-01$\pm$1.42E-01$-$  & 2.80E-01$\pm$3.76E-02 \\
 \cline{2-7}
\multicolumn{1}{c|}{} & $cf_{13}$& 2.19E-01$\pm$3.15E-02$\approx$ & 2.20E-01$\pm$3.25E-02$\approx$ & 2.18E-01$\pm$3.79E-02$\approx$ & 2.32E-01$\pm$4.46E-02$-$   & 2.21E-01$\pm$3.64E-02 \\
\cline{2-7}
\multicolumn{1}{c|}{} & $cf_{14}$& 2.27E-01$\pm$3.29E-02$\approx$ & 2.41E-01$\pm$3.18E-02$-$ & 2.34E-01$\pm$3.28E-02$-$ & 2.15E-01$\pm$6.23E-02$\approx$  & 2.24E-01$\pm$3.04E-02 \\
\cline{2-7}
\multicolumn{1}{c|}{} & $cf_{15}$ & 3.11E+00$\pm$4.27E-01$\approx$ & 3.20E+00$\pm$4.55E-01$\approx$ & 3.22E+00$\pm$3.49E-01$\approx$ & 4.00E+00$\pm$6.87E-01$-$   & 3.16E+00$\pm$3.76E-01 \\
\cline{2-7}
\multicolumn{1}{c|}{} & $cf_{16}$ & 9.52E+00$\pm$3.68E-01$-$ & 9.30E+00$\pm$4.61E-01$\approx$ & 9.68E+00$\pm$3.43E-01$-$ & 1.08E+01$\pm$5.39E-01$-$   & 9.41E+00$\pm$4.33E-01 \\
\cline{1-7}
\multicolumn{1}{c|}{\multirow{6}{*}{\tabincell{c}{Hybrid\\Functions}} }& $cf_{17}$  & 4.11E+02$\pm$1.56E+02$\approx$ & 1.91E+04$\pm$1.08E+05$-$ & 3.87E+02$\pm$1.94E+02$+$ & 5.52E+02$\pm$2.33E+02$-$   & 4.11E+02$\pm$1.56E+02 \\
\cline{2-7}
\multicolumn{1}{c|}{} & $cf_{18}$ & 9.21E+01$\pm$2.98E+01$-$ & 1.14E+02$\pm$1.97E+02$-$ & 2.04E+01$\pm$1.42E+01$+$ & 8.60E+01$\pm$2.63E+01$-$   & 2.94E+01$\pm$1.95E+01 \\
\cline{2-7}
\multicolumn{1}{c|}{} & $cf_{19}$ & 4.79E+00$\pm$8.02E-01$-$ & 4.48E+00$\pm$7.56E-01$\approx$ & 4.48E+00$\pm$7.72E-01$\approx$ & 5.89E+00$\pm$1.11E+00$-$   & 4.48E+00$\pm$7.83E-01 \\
\cline{2-7}
\multicolumn{1}{c|}{} & $cf_{20}$ & 2.29E+01$\pm$1.11E+01$-$ & 3.11E+03$\pm$3.01E+03$-$  & 1.18E+01$\pm$9.27E+00$\approx$ & 4.69E+01$\pm$2.02E+01$-$  & 1.22E+01$\pm$4.65E+00 \\
\cline{2-7}
\multicolumn{1}{c|}{} & $cf_{21}$& 3.66E+02$\pm$1.99E+02$-$ & 1.33E+04$\pm$4.12E+04$-$ & 1.60E+02$\pm$8.74E+01$-$ & 4.51E+02$\pm$1.34E+02$-$   & 1.40E+02$\pm$1.00E+02 \\
\cline{2-7}
\multicolumn{1}{c|}{} & $cf_{22}$ & 1.30E+02$\pm$6.55E+01$-$ & 1.44E+02$\pm$7.74E+01$-$ & 9.76E+01$\pm$6.22E+01$+$ & 1.48E+02$\pm$7.06E+01$-$   & 1.12E+02$\pm$7.46E+01 \\
\cline{1-7}
\multicolumn{1}{c|}{\multirow{8}{*}{\tabincell{c}{Composition\\Functions }}} & $cf_{23}$ & 3.15E+02$\pm$4.01E-13$\approx$ & 3.15E+02$\pm$4.01E-13$\approx$ & 3.15E+02$\pm$4.01E-13$\approx$ & 3.15E+02$\pm$4.01E-13$\approx$  & 3.15E+02$\pm$3.59E-13 \\
\cline{2-7}
\multicolumn{1}{c|}{} & $cf_{24}$ & 2.24E+02$\pm$1.81E+00$\approx$ & 2.25E+02$\pm$3.60E+00$\approx$& 2.24E+02$\pm$2.59E+00$\approx$ & 2.27E+02$\pm$4.62E+00$\approx$   & 2.24E+02$\pm$1.85E+00 \\
\cline{2-7}
\multicolumn{1}{c|}{} & $cf_{25}$ & 2.03E+02$\pm$5.71E-01$\approx$ & 2.03E+02$\pm$1.13E+00$\approx$& 2.02E+02$\pm$2.99E-01$\approx$ & 2.03E+02$\pm$5.66E-01$\approx$   & 2.03E+02$\pm$4.16E-01 \\
\cline{2-7}
\multicolumn{1}{c|}{} & $cf_{26}$ & 1.06E+02$\pm$2.37E+01$-$ & 1.02E+02$\pm$1.39E+01$\approx$ & 1.00E+02$\pm$3.50E-02$\approx$ & 1.00E+02$\pm$5.39E-02$\approx$   & 1.00E+02$\pm$4.27E-02 \\
\cline{2-7}
\multicolumn{1}{c|}{} & $cf_{27}$ & 3.43E+02$\pm$4.75E+01$\approx$ & 3.35E+02$\pm$4.68E+01$\approx$ & 3.49E+02$\pm$4.80E+01$-$  & 3.82E+02$\pm$4.18E+01$-$ & 3.34E+02$\pm$4.61E+01 \\
\cline{2-7}
\multicolumn{1}{c|}{} & $cf_{28}$ & 8.01E+02$\pm$4.06E+01$\approx$ & 7.96E+02$\pm$4.63E+01$\approx$ & 8.06E+02$\pm$3.56E+01$-$ & 8.58E+02$\pm$6.96E+01$-$   & 7.96E+02$\pm$4.57E+01 \\
\cline{2-7}
\multicolumn{1}{c|}{} & $cf_{29}$ & 1.80E+05$\pm$1.28E+06$-$ & 8.28E+02$\pm$3.27E+02$-$ & 7.07E+02$\pm$8.30E+02$-$ & 4.60E+05$\pm$2.31E+06$-$  & 6.12E+02$\pm$2.00E+02 \\
\cline{2-7}
\multicolumn{1}{c|}{} & $cf_{30}$ & 1.53E+03$\pm$7.69E+02$-$  & 1.66E+03$\pm$7.61E+02$-$ & 1.13E+03$\pm$4.87E+02$-$ & 1.27E+03$\pm$6.65E+02$-$ & 1.02E+03$\pm$4.21E+02 \\
\hline
\multicolumn{2}{c|}{$+$}        &  0 & 0  & 3  & 0\\  \cline{1-6}
\multicolumn{2}{c|}{$-$}        & 11 & 13 & 11 &21\\  \cline{1-6}
\multicolumn{2}{c|}{ $\approx$} & 19 & 17 & 16 &9 \\  \cline{1-6}
\end{tabular}

\end{sidewaystable*}

We are interested in identifying the benefit of two distinguished components of ACoS: the additional archiving mechanism and  the probability vector $\vec{p}$. To this end, we still selected JADE as the instance algorithm and two groups of experiments were carried out. In the first group, the archiving mechanism was eliminated and the offspring in the current generation played the role of the archive $\textbf{A}$ in ACoS accordingly, while the other parts were kept untouched. This compared method is denoted as nonAr-ACoS-JADE. With respect to the second group, instead of adaptive tuning, $\vec{p}$ was fixed during the evolution. We tested three different values for each element of $\vec{p}$: 0, 0.5, and 1, and these three values represent different conditions, i.e., only the original coordinate system is used, the Eigen and original coordinate systems have an equal probability to be selected, and only the Eigen coordinate system is utilized, respectively. It is evident that the first condition is equivalent to the original JADE. These compared methods are named as JADE, half-ACoS-JADE, and Eig-ACoS-JADE, respectively.

We conducted the experiments on $cf_{1}$-$cf_{30}$ with 30D. Experimental results are presented in Table VI, where ``$+$'', ``$-$'', and ``$\approx$'' denote that the performance of the corresponding algorithm is better than, worse than, and similar to that of  ACoS-JADE, respectively. From Table VI, ACoS-JADE performs the best among the five compared methods. Compared with nonAr-ACoS-JADE, ACoS-JADE is significantly better on 11 test functions and does not lose on any test functions. Although ACoS-JADE and nonAr-ACoS-JADE achieve comparable performance  on the unimodal functions, ACoS-JADE outperforms nonAr-ACoS-JADE  on more complex functions (i.e., Simple multimodal functions, hybrid functions, and composition functions). The reason is probably that the additional archiving mechanism preserves the offspring not only in the current generation but also in the past several generations, thus providing sufficient information to estimate a more reliable Eigen coordinate system in complex environments. Compared with JADE, half-ACoS-JADE, and Eig-ACoS-JADE, ACoS-JADE produces better results on 13, 11 and 20 test functions, respectively; while the three competitors cannot outperform ACoS-JADE on more than three test functions. This phenomenon suggests that the updating of $\vec{p}$ in our framework has the capability to provide a more proper coordinate system. It is noteworthy that ACoS-JADE and half-ACoS-JADE have an advantage over JADE and Eig-ACoS-JADE, which again verifies the effectiveness of combining both the Eigen and original coordinate systems together.

From the above discussion, one can conclude that both the additional archiving mechanism and the probability vector $\vec{p}$ play very important roles in ACoS. The former is beneficial to estimate a more reliable Eigen coordinate system, and the latter enables each individual to select a more appropriate coordinate system. In addition, the utilization of both the Eigen and original coordinate systems is quite necessary in the design of an EA.

\begin{figure*} [!t]
    \begin{center}
      \subfigure[$cf_{1}$ with 30D]{\includegraphics[width=4cm]{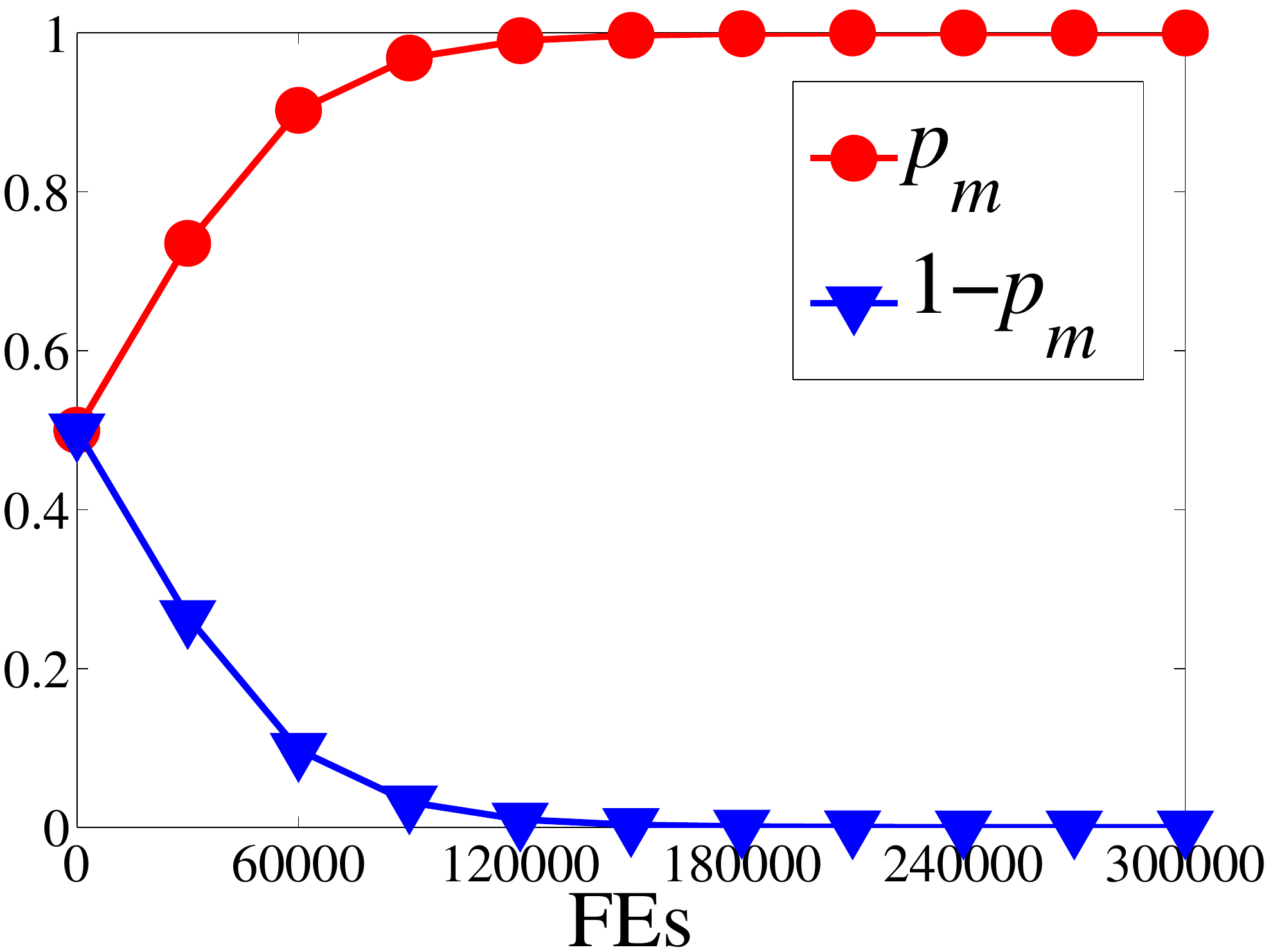}}
        \subfigure[$cf_{10}$ with 30D]{\includegraphics[width=4cm]{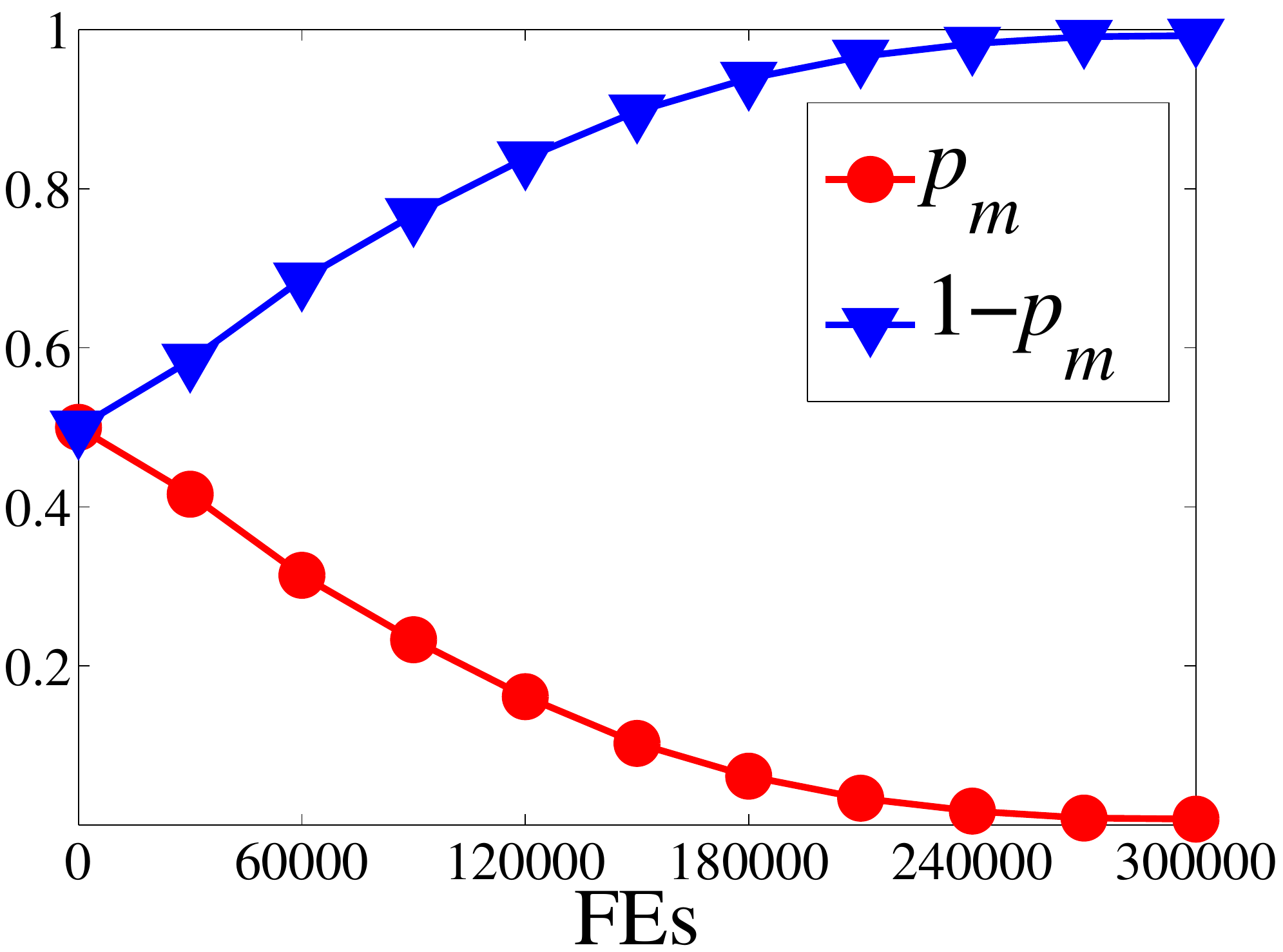}}
         \subfigure[$cf_{23}$ with 30D]{\includegraphics[width=4cm]{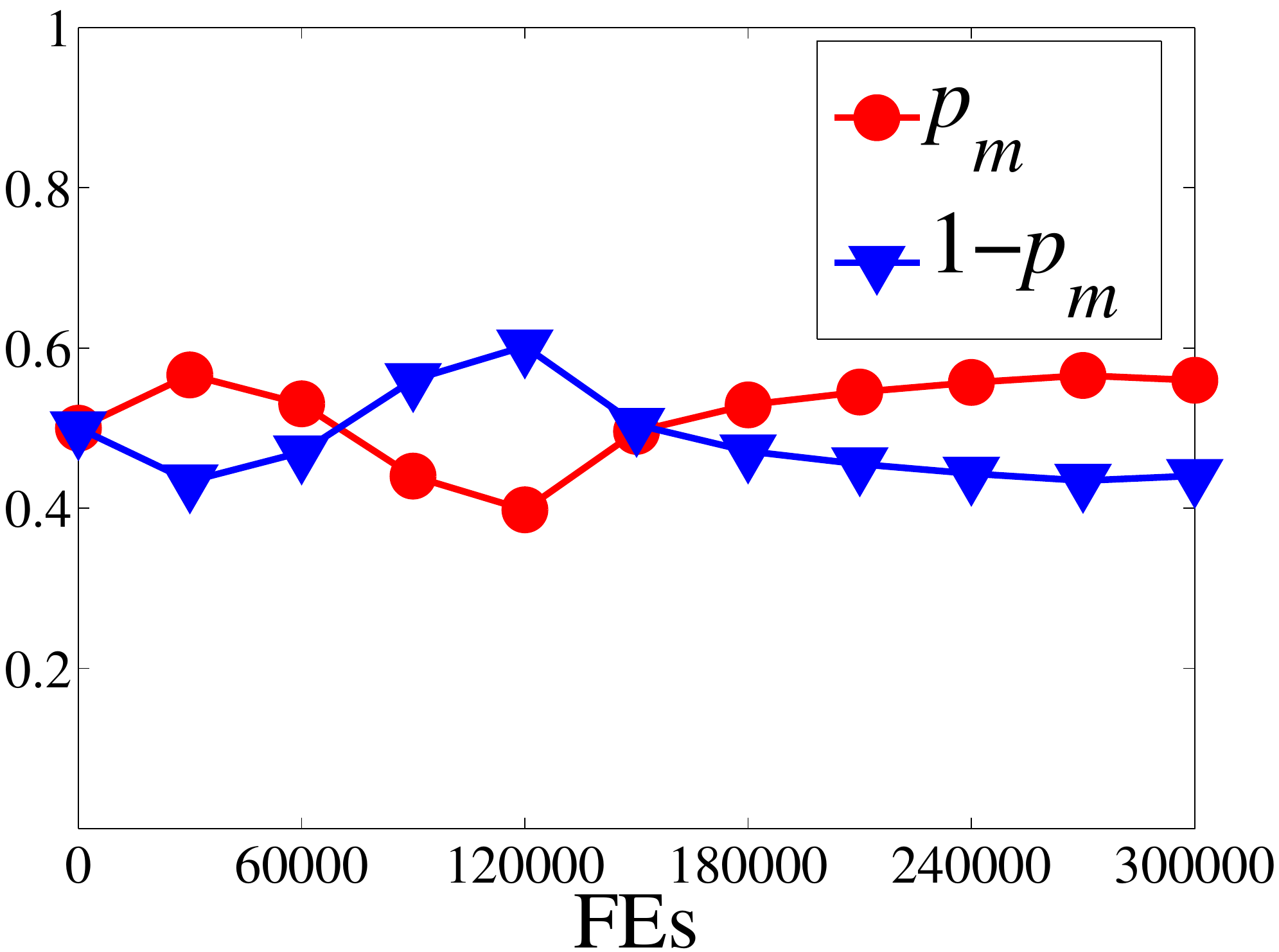}}
       \caption{The evolution of the average values of $p_{m}$ and $1-p_{m}$ in ACoS-JADE in the optimization of $cf_{1}$, $cf_{10}$ and $cf_{23}$  }
    \end{center} \label{fig:pmonitor}
\end{figure*}

\subsection {Evolution of the probability vector in ACoS}
Since the probability vector $\vec{p}=(p_{1},p_{2},...p_{NP})$ determines the selection ratio of each coordinate system for each individual, one may be interested in investigating the dynamic changes of $\vec{p}$ during the evolutionary search. For this purpose, the mean value of $\vec{p}$, referred as $p_{m}={1 \over {NP}}\sum\nolimits_{i = 1}^{NP} {{p_i}}$, is monitored in this subsection.

We still chose JADE as the instance algorithm and tested ACoS-JADE on three test functions with 30D from IEEE CEC2014: the unimodal function $cf_{1}$, the simple multimodal function $cf_{10}$, and the composite function $cf_{23}$. These three different kinds of test functions aim to provide a comprehensive study on the changes of $\vec{p}$. To visualize the results, Fig. 4 plots the evolution of the average values of $p_{m}$ and $1-p_{m}$ over 51 independent runs.

As shown in Fig. 4, there are three different types of curves. In the first type (see Fig. 4(a)), the Eigen coordinate system has a larger probability to be selected than the original coordinate system. Nevertheless, in the second type (see Fig. 4 (b)), the situation is opposite. For the third type (see Fig. 4(c)), these two coordinate systems have the similar probability to be chosen over the course of evolution. It can be seen from Table VI that Eig-ACoS-JADE outperforms JADE on $cf_{1}$, JADE surpasses Eig-ACoS-JADE on $cf_{10}$, and Eig-ACoS-JADE and JADE reach the similar performance on $cf_{23}$, which implies that the Eigen coordinate system is more appropriate for $cf_{1}$, the original coordinate system is a better choice for $cf_{10}$, and these two coordinate systems are both important for $cf_{23}$, respectively. Interestingly, the changes of $p_{m}$ and $1-p_{m}$ in Fig. 4 are consistent with the above analysis, which indicates that ACoS is able to adapt $\vec{p}$ to a reasonable value to match different function landscapes. In summary, the following concludes can be made: 1) there does not exist a one-size-fits-all coordinate system, and 2) our proposed framework can effectively select the appropriate coordinate system for different optimization problems.

\section {Conclusion}\label{sec:Conclusion}
An adaptive framework for tuning the coordinate systems in EAs referred as ACoS has been proposed in this paper. ACoS provides a simple yet efficient approach to establish the Eigen coordinate system via an additional archiving mechanism and the rank-$\mu$-update strategy. Thereafter, it adopts a probability vector $\vec{p}$, which is adaptively updated by making use of the collected information from the offspring, to select an appropriate coordinate system between the Eigen and original coordinate systems for an EA. This paper also presents a new point of view toward how to transform an evolutionary operator in the original coordinate system into the corresponding evolutionary operator in the Eigen coordinate system. We have applied ACoS to two of the most popular EA paradigms, namely PSO and DE, for solving test functions with 30D and 50D from IEEE CEC2014. Simulation results demonstrate that ACoS is capable of significantly enhancing the performance of both PSO and DE. Comparing with some other Eigen coordinate system based methods, ACoS also achieves quite promising results. In the future, we will apply ACoS to improve the performance of other EA paradigms.

The Matlab source code of ACoS can be obtained from the authors upon request.
\bibliographystyle{plain}
\bibliography{ACoS}

\end{document}